\documentclass[a4paper,fleqn]{cas-dc}
\usepackage{lineno,hyperref}
\modulolinenumbers[5]

\usepackage{amsmath,amsfonts}
\usepackage{algorithmic}
\usepackage{algorithm}
\usepackage{array}
\usepackage[caption=false,font=normalsize,labelfont=sf,textfont=sf]{subfig}
\usepackage{textcomp}
\usepackage{stfloats}
\usepackage{url}
\usepackage{graphicx}
\usepackage{verbatim}
\usepackage{bm}
\usepackage{cases}
\usepackage{multirow}
\usepackage{multicol}
\usepackage{booktabs}
\usepackage{chngcntr}
\usepackage{pifont}

\usepackage[authoryear]{natbib}

\def\tsc#1{\csdef{#1}{\textsc{\lowercase{#1}}\xspace}}
\tsc{WGM}
\tsc{QE}

\begin{document}
\let\WriteBookmarks\relax
\def\floatpagepagefraction{1}
\def\textpagefraction{.001}

\shorttitle{MMOTU}    

\shortauthors{Qi Zhao, Shuchang et al.}   

\title [mode = title]{MMOTU: A Multi-Modality Ovarian Tumor Ultrasound Image Dataset for Unsupervised Cross-Domain Semantic Segmentation}  



%

\author[1]{Qi Zhao}[type=author]


\fnmark[1]

\ead{zhaoqi@buaa.edu.cn}



\affiliation[1]{organization={Department of Electronics and Information Engineering, Beihang University},
            addressline={Xueyuan Road No.37, Haidian district}, 
            city={Beijing},
            postcode={100191}, 
            country={China}}

\affiliation[2]{organization={Department of Gynecology and Obstetrics, Beijing Shijitan Hospital, Capital Medical University},
            addressline={Tieyi Road No.10, Haidian district}, 
            city={Beijing},
            postcode={100038}, 
            country={China}}

\affiliation[3]{organization={University of Liverpool},
            addressline={Foundation Building,
Brownlow Hill,}, 
            city={Liverpool},
            postcode={L693BX}, 
            state={},
            country={UK}}

\author[1]{Shuchang Lyu}[type=author, orcid=0000-0001-9769-7083]

\fnmark[1]

\ead{lyushuchang@buaa.edu.cn}



\author[2]{Wenpei Bai}[type=author]
\fnmark[1]
\ead{baiwp@bjsjth.cn}
\author[1]{Linghan Cai}[type=author]
\fnmark[1]
\ead{cailh@buaa.edu.cn}
\author[1]{Binghao Liu}[type=author]
\ead{liubinghao@buaa.edu.cn}
\author[3]{Guangliang Cheng}[type=author]
\ead{Guangliang.Cheng@liverpool.ac.uk}
\author[2]{Meijing Wu}[type=author]
\ead{nancywu0429@foxmail.com}
\author[2]{Xiubo Sang}[type=author]
\ead{sangxiubo3506@bjsjth.cn}
\author[2]{Min Yang}[type=author]
\cormark[2]
\ead{yangmin@bjsjth.cn}
\author[1]{Lijiang Chen}[type=author]
\cormark[1]
\ead{chenlijiang@buaa.edu.cn}

\cortext[1]{Primary Corresponding author}
\cortext[2]{Secondary Corresponding author}

\fntext[1]{Contribute Equally.}


\begin{abstract}
Ovarian cancer is one of the most harmful gynecological diseases. Detecting ovarian tumors in early stage with computer-aided techniques can efficiently decrease the mortality rate. With the improvement of medical treatment standard, ultrasound images are widely applied in clinical treatment. However, recent notable methods mainly focus on single-modality ultrasound ovarian tumor segmentation or recognition, which means there still lacks researches on exploring the representation capability of multi-modality ultrasound ovarian tumor images. To solve this problem, we propose a Multi-Modality Ovarian Tumor Ultrasound (MMOTU) image dataset containing 1469 2d ultrasound images and 170 contrast enhanced ultrasonography (CEUS) images with pixel-wise and global-wise annotations. Based on MMOTU, we mainly focus on unsupervised cross-domain semantic segmentation task. To solve the domain shift problem, we propose a feature alignment based architecture named Dual-Scheme Domain-Selected Network (DS$^2$Net). Specifically, we first design source-encoder and target-encoder to extract two-style features of source and target images. Then, we propose Domain-Distinct Selected Module (DDSM) and Domain-Universal Selected Module (DUSM) to represent the distinct and universal features in two styles (source-style or target-style). Finally, we fuse these two kinds of features and feed them into the source-decoder and target-decoder to generate final predictions. Extensive comparison experiments and analysis on MMOTU image dataset show that DS$^2$Net can boost the segmentation performance for bidirectional cross-domain adaptation of 2d ultrasound and CEUS images. Our proposed dataset and code are all available at \url{https://github.com/cv516Buaa/MMOTU_DS2Net}.
\end{abstract}



\maketitle

\section{Introduction}
\label{sec:introduction}
\par Ovarian cancer is one of the most mortal gynecological diseases, which ranks $8^{th}$ among all the gynecological cancers~\cite{2021Cancer}. Diagnosing and detecting ovarian tumors in the early stage can significantly decrease the mortality rate. At present, the common-used screening techniques are two-dimensional (2d) ultrasound scanning, contrast enhanced ultrasonography (CEUS), Computed Tomography (CT) and Magnetic Resonance Imaging (MRI). Among them, 2d ultrasound scanning is the most widely-applied technique, because it is more convenient and has less impact on human body.
\par Recently, several methods focus on computer-aided detecting and diagnosing ovarian tumors~\cite{DL_OTC, OT_cls, ovarySeg_MIA, HASA, CR-UNet, CTransCNN}. Despite their notable works, there still exists the following two main weaknesses. First, recent methods only focus on single-modality segmentation and recognition (mainly on 2d ultrasound images). There still lacks researches on exploring the representation potential of multi-modality ultrasound images because of lacking standard datasets. Second, even though many notable methods~\cite{AdapSegNet, CyCADA, cyclegan, SIFA, DSFN} make huge progress on cross-domain adaptation, there still lacks solution on tackling cross-domain segmentation among multi-modality ultrasound images.
\begin{figure}[tb]
\begin{center}
   \includegraphics[width=1.0\linewidth]{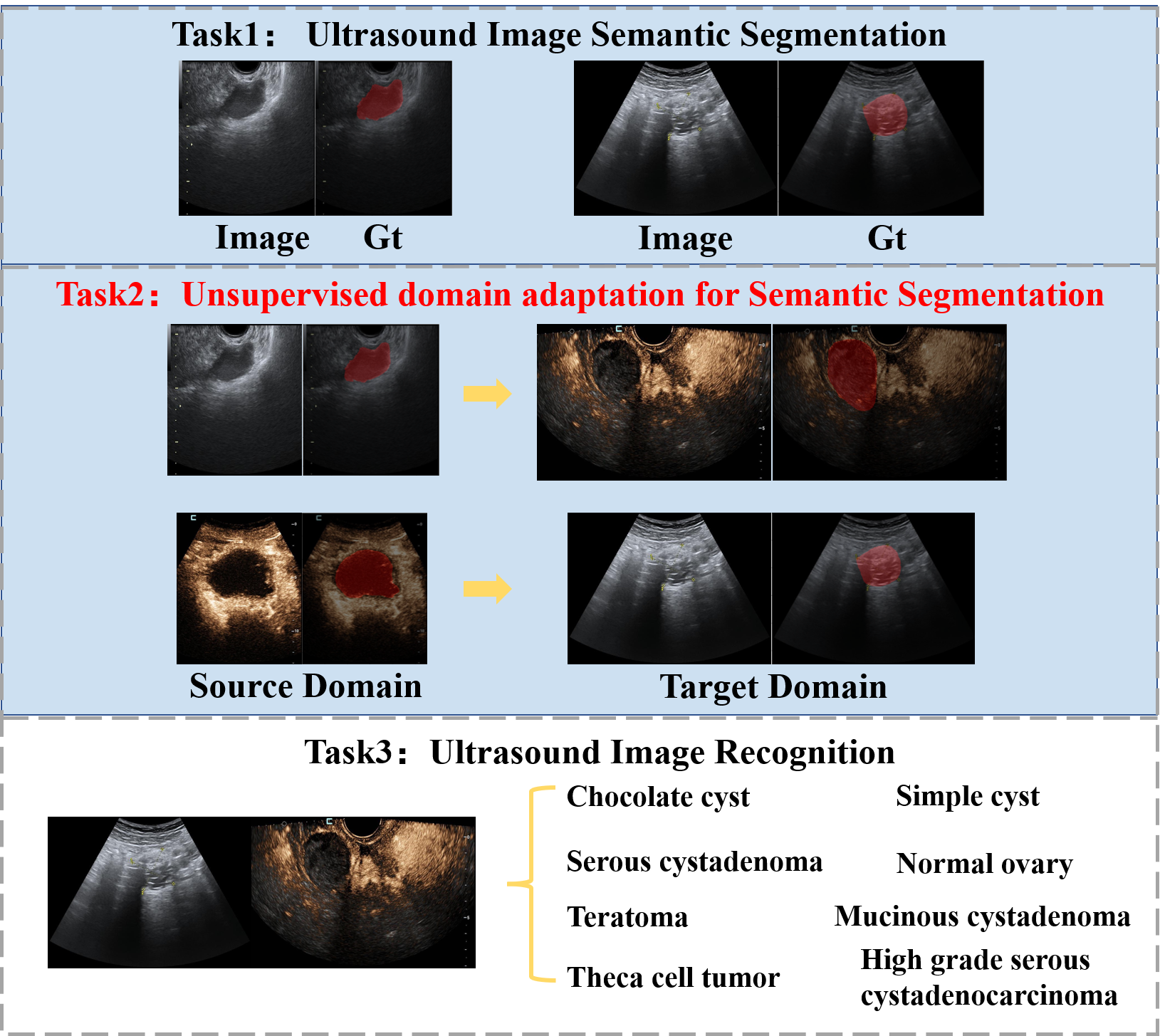}
\end{center}
   \caption{Task description on MMOTU image dataset. \textbf{Task1:} single-modality semantic segmentation. \textbf{Task2:} bidirectional unsupervised domain adaptation between OTU\_2d and OTU\_CEUS for semantic segmentation. \textbf{Task3:} single-modality image recognition. Here, Task2 is our main research focus. Task1 is the prior research of Task2. Task3 is an independent task, which is also meaningful in clinical treatment.}
\label{Fig1}
\end{figure}
\par To solve the above-mentioned two weaknesses, we first construct a Multi-Modality Ovarian Tumor Ultrasound (MMOTU) image dataset. MMOTU image dataset consists of two sub-sets with two modalities, which respectively contain 1469 2d ultrasound images (OTU\_2d) and 170 CEUS images (OTU\_CEUS). On these two sub-sets, we provide pixel-wise semantic annotations and global-wise category annotations. Then, we explore the cross-domain representation potential on MMOTU image dataset. As shown in Fig.\ref{Fig1}, we tackle three tasks on MMOTU image dataset. The second task (unsupervised domain adaptation based semantic segmentation) is our main research focus, which aims to boost the segmentation performance on target samples with only source annotated samples for training. The first task (semantic segmentation) is the prior research of second task. The third task (recognition task) is also meaningful for clinical treatment.
\par Recently, deep convolutional neural networks (DCNNs) have shown great success on recognition~\cite{Chi_etal, Wang_etal, CancerImgCLS, DSI-Net, mmriCLS, DL_OTC} and semantic segmentation~\cite{UNet, TransUNet, UNet++, Transfuse} of medical images. These two tasks play an important role on computer-aided clinical diagnosis. Based on MMOTU image dataset, we first tackle single-modality semantic segmentation task, which aims to find the lesion region. In clinical treatment, this task can apply on computing the measurements. We typically hope to report tumor size using segmentation result as prior knowledge. This task also provides evidence for recognition results, which can help inexperienced young doctors to make more accurate diagnosis with lesion region information. In this paper, we provide four types of notable baseline architectures in semantic segmentation, which are CNN-based ``Encoder-Decoder'', transformer-based ``Encoder-Decoder'', U-shape networks and spatial-context based two-branch networks. Specifically, CNN-based ``Encoder-Decoder'' is a classical segmentation architecture. Transformer-based ``Encoder-Decoder'' integrates transformers into segmentors to enhance the performance with self-attention mechanism. U-shape networks are widely used on medical image segmentation, such as the widely applied U-Net~\cite{UNet}. Spatial-context based two-branch architecture is another efficient choice. It designs two branches to respectively capture low-level high-resolution features and high-level context features.
\par Based on baseline segmentors, we then tackle the unsupervised domain adaptation based semantic segmentation task. Recent UDA methods solve the domain shift problem mainly by image alignment~\cite{DEUDA, SCUDA, SIFA, DSFN} and feature alignment~\cite{AdapSegNet, SSF-DAN, PBUDA, EGUDA}. In this paper, we propose a feature alignment based method to solve domain shift problem. The motivations can be summarized in the following points: (1) Compared to image alignment based methods, feature alignment based methods always require less computation resource in both training and testing phases. (2) Image alignment based methods show strong power when appearance of source and target images have dramatic difference. Intuitively, the main appearance shift between 2d ultrasound and CEUS images is the color (Fig.\ref{Fig1}), which means this appearance shift may not cause heavy feature-shift especially on high-level features. (3) Based on the baseline segmentors trained with source images, we directly apply them on target images and find that the segmentation performance on source-trained model does not decrease dramatically compared to the performance on target-trained model. This observation encourages us to explore the potential of feature alignment based method. (4) Image alignment methods mainly utilize ``pixel-to-pixel'' image translation technique to transform target image into source-style. The inaccuracy and distortion of source-style target image highly harm the segmentation performance. (5) image alignment method depends on generative technique (``pixel-to-pixel" image translation). Compared to feature alignment method, it requires much larger training samples. Obviously, MMOTU image dataset is not that large (OTU\_2d and OTU\_CEUS respectively contains 1469 and 170 samples), which will lead to unstable optimization and low translation quality. As a whole, feature alignment method is more suitable for the tasks on our proposed dataset.
\par In this paper, we propose a feature alignment based architecture, which is Dual-Scheme Domain-Selected Network (DS$^2$Net). Specifically, we first design two encoders ($E_{s}$ and $E_{t}$) to map source and target images into two-style features (i.e. $E_{s}$ will respectively map source and target images into source-style source features and source-style target features. Similar for $E_{t}$). Then, we assume that two-style features are not decoupled. It means source-style features may contain some target-style information and vice versa. Based on this assumption, we propose Domain-Distinct-Selected Module (DDSM). With DDSM, we can decouple the source-style (target-style) features into purer source-style (target-style) features and target-style (source-style) features. Obviously, some information may work on both sides or not work on any side. Therefore, we further propose Domain-Universal-Selected-Module (DUSM). With DUSM, we further find the universal information of source/target-style features. Finally, we respectively feed the source-style and target-style features into two decoder-heads ($H_{s}$ and $H_{t}$) for final predictions. During optimization, besides supervision from source annotations, we apply adversarial learning to combat the domain shift on the two-style features output from $E_{s}$ and $E_{t}$. This design ensures that two-style features can be aligned into same latent space. To prove the effectiveness of DS$^2$Net, we conduct comparison experiments with recent notable feature alignment based domain adaptation methods~\cite{AdapSegNet, EGUDA} on MMOTU image dataset. The experiment results are encouraging.
\par In summary, the main contributions are listed as follows:
\begin{itemize}
\item We construct MMOTU image dataset containing 2d ultrasound and CEUS images. For every images in dataset, we provide pixel-wise semantic annotations and global-wise category annotations.
\item To the best of our knowledge, we first propose a method (DS$^2$Net) to tackle the cross-domain ovarian tumor segmentation between 2d ultrasound and CEUS images. With this method, we provide an insight on detecting ovarian tumors on multi-modality ultrasound images.
\item DS$^2$Net utilizes domain selected modules (DDSM and DUSM) to extract domain-distinct and domain-universal features. Essentially, we provide an insight on using feature decoupling technique to solve domain shift problem.
\end{itemize}
\section{Related Work} 
\label{sec:relatedwork}
\subsection{Computer-Aided Methods on Ovarian Ultrasound Image Datasets}
\par To develop computer-aided methods on medical treatment of ovarian diseases, many methods construct ultrasound image datasets and use DCNNs to tackle the image recognition task. ~\cite{DL_OTC} construct a 2d ultrasound image dataset with three types of ovarian tumors (Benign, Borderline and Malignant). This dataset is used for ovarian tumor classification. Followed ~\cite{DL_OTC}, ~\cite{Wang_etal} collect annotated samples of serous ovarian tumors (SOTs) and then apply DCNNs to categorize SOTs into Benign, Borderline and Malignant. 
\par Besides recognition task, many notable methods are proposed to tackle segmentation task on 2d ultrasound ovarian images. These methods mainly aim at characterizing the ovarian structure.~\cite{CR-UNet} propose CR-UNet, which integrates the spatial recurrent neural network (RNN) into a plain U-Net to segment the ovary and follicles.~\cite{e2e_OTSeg} propose fCNN to automatically characterize ovarian structures. Both CR-UNet and fCNN are trained on their collected 2d ultrasound ovarian image datasets with pixel-wise annotations. Recently, several methods have been proposed to characterize the ovarian structure.~\cite{S-Net} propose S-Net to simultaneously segment ovary and follicles in 3d-ultrasound volumes.~\cite{ovarySeg_MIA} propose C-Rend (contrastive rending) on ovary and follicles segmentation and further apply it into a semi-supervised learning framework. Both S-Net and C-Rend are trained on their collected 3d-ultrasound ovarian volume datasets.
\subsection{Semantic Segmentation on Medical Images}
\par Semantic segmentation has become an widely applied medical image processing technique on many clinical applications, such as organ structure characterization, lesion area detection and cell segmentation. In this field, U-Net~\cite{UNet} is one of the most famous methods, which creates shortcuts between encoder and decoder to capture contextual and precise localization information. Followed U-Net, U-Net++~\cite{UNet++} connects encoder and decoder using a series of nested dense convolutional blocks. Compared to U-Net, this design further bridges the semantic gap between feature maps from encoder to decoder. UcUNet~\cite{UcUNet} designs an efficient U-shaped convolution block to increase the network depth with fewer parameters and efficiently ignore invalid features while fusing shallow and deep features. In recent years, attention mechanism substantially promotes semantic segmentation.~\cite{att_U-Net} propose an attention U-Net. Before fusing the feature maps from encoder to decoder, attention U-Net inserts an attention gate to control the spatial-wise feature importance. Besides applying spatial-wise attention mechanism, ~\cite{FEDNet} propose FED-Net, which uses a channel-wise attention mechanism to improve the performance of liver lesion segmentation. There are also many works~\cite{FocusNet, 3DQ, NL_UNet, TransMIS} aiming at mixing spatial-wise and channel-wise attention mechanisms. 
\par Transformer is a cutting-edge structure utilizing self-attention mechanism to capture important region of images. On medical image semantic segmentation, TransFuse~\cite{Transfuse} and TransUNet~\cite{TransUNet} both employ ViT (Vision Transformer)~\cite{ViT} as encoder for feature extraction. Swin-UNet~\cite{swinUnet} applies Swin-Transformer~\cite{Swin} on both encoder and decoder to explore the learning potential on long-range semantic information. SegTran~\cite{segTran} further proposes a squeeze-and-expansion transformer for more diversified representations. MAXFormer~\cite{MAXFormer} utilizes an efficient parallel local–global transformer module to enhance accuracy of medical image segmentation for clinical applications.
\subsection{Unsupervised Domain Adaptation on Medical Images}
\par UDA mainly aims to alleviate the domain shift problem when applying a source-trained model on target domain data. There are mainly two types of methods on UDA based semantic segmentation task, which are image alignment and feature alignment based methods. Image alignment based methods always adopt image translation~\cite{cyclegan, pix2pix} to align between source and target images by appearance transformation.~\cite{SIFA} and \cite{DEUDA} propose generators for source-to-target transformation and then optimize the segmentor by target-style source images with corresponding annotations.~\cite{DSFN} and \cite{SCUDA} further add target-to-source generators besides source-to-target generators to bridge source and target domain with a balanced flow. Feature alignment based methods solve the domain shift problem through exploring domain invariant features.~\cite{EGUDA} propose feature and entropy map discriminators to explore the representation potential of domain invariant features.~\cite{PBUDA} propose pOSAL to segment the optic disc and optic cup from different fundus image datasets in a joint manner. pOSAL exploits UDA with feature alignment mechanism to ease the domain shift.~\cite{PnP-AdaNet} propose PnP-AdaNet to tackle domain shift problem by aligning source and target features. On low-level and high-level latent space, it applies adversarial loss for feature-level adaptation. Moreover, some other notable feature alignment based methods, such as AdapSegNet~\cite{AdapSegNet} and SSF-DAN~\cite{SSF-DAN} both achieve great success on both natural scene images and medical images.
\par There are also many works focusing on disentangling features in unsupervised cross-domain medical image semantic segmentation task.~\cite{UDA-BTS} and \cite{DADR} both propose a cross-domain semantic segmentation framework, which decomposes cross-domain images into a domain-invariant content space and domain-specific style space by feature disentanglement module. Both methods achieve outstanding robustness and the generalized capability on cross-domain learning and robust adaptation. Similar to \cite{DADR, UDA-BTS}, \cite{DDFSeg} propose an image alignment based method (DDFSeg) to tackle cross-modality cardiac image segmentation task. Besides extracting domain-invariant features (DIFs), DDFSeg further considers the domain-specific features (DSFs) as complementary features for image translation. With better synthetic source-style target images, the source-trained segmentation network can predict better. To overcome inherent misregistration and disparity in signal intensity on cross-domain images, \cite{DAFNet} propose DAFNet to provide improved segmentation accuracy by leveraging information of images of other modalities under unsupervised configuration. DAFNet utilizes disentangled decomposition to disentangle images into semantic anatomy factors and modality factors to map cross-domain images into a smooth latent space. \cite{MML} and \cite{HECKTOR} investigate the feasibility and effectiveness of promoting single-modality medical segmentation with aid of multi-modality information.
\begin{table}
\caption{The data distribution of MMOTU image dataset.}
\begin{center}
\setlength{\tabcolsep}{1mm}{
\begin{tabular}{cccccc}
 \cmidrule(r){1-6}
  \multirow{2}{*}{Data type} & \multirow{2}{*}{categories} & \multicolumn{2}{c}{Training set} & \multicolumn{2}{c}{Testing set}
  \\ \cmidrule(r){3-6}
  {} & {} & samples & patients & samples & patients 
  \\ \cmidrule(r){1-6}
  OTU\_2d & 8 & 1000 & 171 & 469 & 76
  \\ \cmidrule(r){1-6}
  OTU\_CEUS & 8 & 70 & 20 & 100 & 27
  \\ \cmidrule(r){1-6}
\end{tabular}}
\end{center}
\label{Tab1}
\end{table}
\begin{figure}[tb]
\begin{center}
   \includegraphics[width=1.0\linewidth]{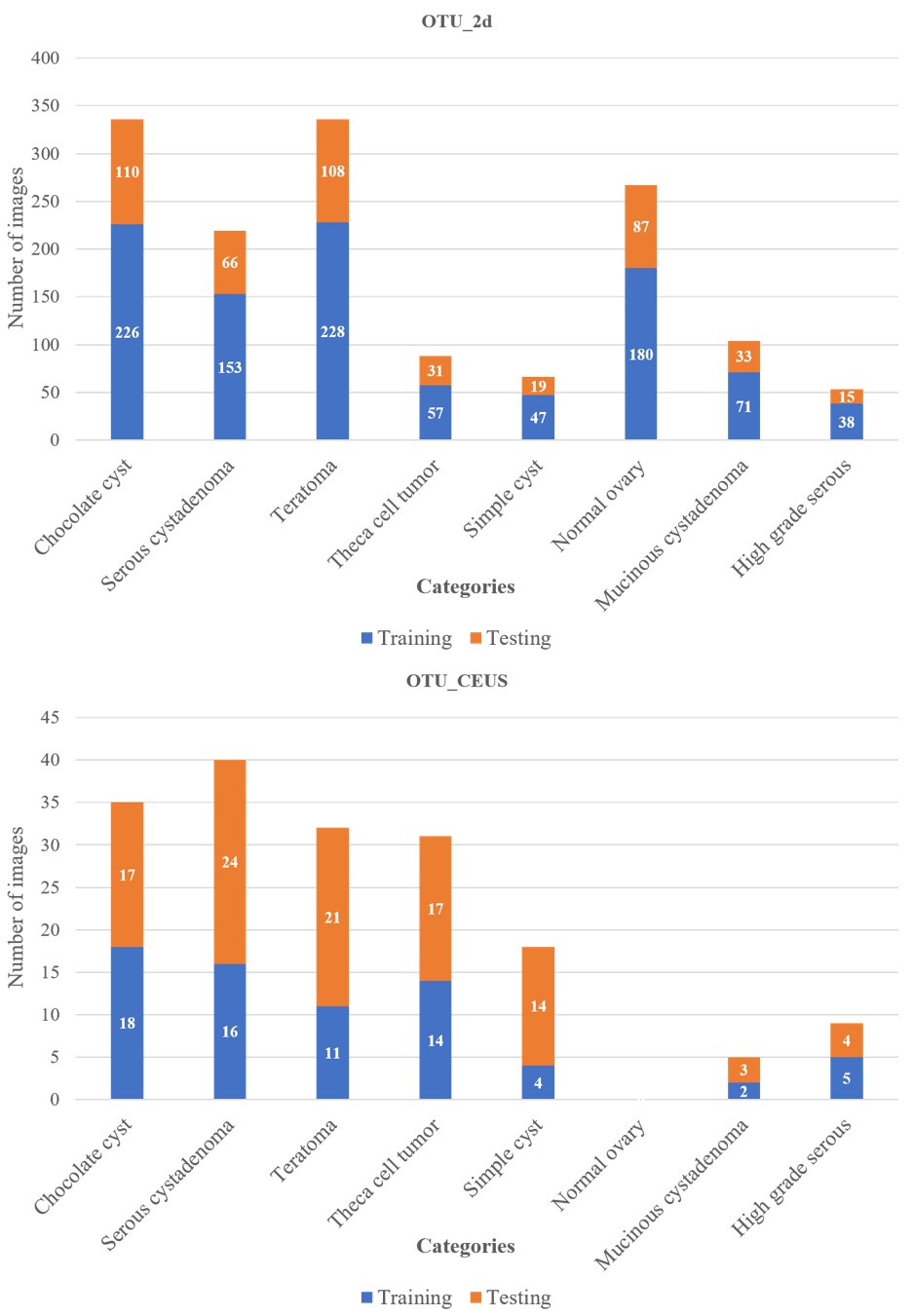}
\end{center}
   \caption{The number of samples containing in each category.}
\label{Fig_2}
\end{figure}
\section{Dataset}
\label{sec:dataset}
\subsection{Establishment}
\par To tackle the cross-domain ovarian tumor segmentation task, we first propose MMOTU image dataset. All images are obtained from Beijing Shijitan Hospital, Capital Medical University. Our dataset contains 1639 ovarian ultrasound images collected from 294 patients. The scanner is Mindray Resona8 ultrasonic diagnostic instrument. Obviously, each patient is collected with multiple scans. During scanning, doctors select some images (slices), which can clearly show the lesion regions. Sometimes, one slice cannot meet the satisfaction of diagnosis, so doctors will select multiple scans. The main difference between those scans is view. Transvaginal ultrasound scanning is a dynamic process, so multiple scans are extracted from different view angles. As shown in Tab.\ref{Tab1}, MMOTU image dataset contains two sub-sets with two modalities, where OTU\_2d and OTU\_CEUS respectively consist of 1469 2d ultrasound images and 170 CEUS images. On two sub-sets, we split them into training and testing sets. Since each patient is collected with multiple scans, we carefully deal with all samples during splitting dataset to make sure that no ``patient-overlapping'' between training and testing sets. The number of patients in training and testing set are shown in Tab.\ref{Tab1}. It is worth noting that, testing set contains more images than training set in OTU\_CEUS, because we hope to guarantee the evaluation quality. When we tackle cross-domain segmentation, more testing images make the results more convincing. Since we will continually collect CEUS images, more samples will be soon added in training set.  
\par MMOTU image dataset has eight typical categories of ovarian tumor (Fig.\ref{Fig1}). Fig.\ref{Fig_2} further shows the sample distribution of each category in detail. From Fig.\ref{Fig_2}, we find the following two points. (1) Samples of each category are unbalanced. This is because some types tumors are more common while some types of tumors are rarer in clinical treatment. (2) In OTU\_CEUS, some categories only contain few samples, which is easy to cause underfitting. Due to historical and practical reasons, most of the CEUS sequences/images are not stored. Even though we can not provide more samples for those few-shot categories, it is a chance for researchers to conduct researches on few-shot ovarian tumor segmentation or classification. In this research, providing an insight on AI-aided medical treatment using CEUS sequences/images is our main aim. In the future, we will pay attention on preserving and continually collecting CEUS images to extend our OTU\_CEUS dataset. (3) ``Normal Ovary'' category has no sample in OTU\_CEUS, because CEUS technology is usually a further examination after 2d ultrasound examination, which is mainly applied to confirm the tumor types (e.g. Benign, Borderline, Malignant or our proposed specific types). It means the CEUS images hardly have ``Normal Ovary'' category. Under this situation, the segmentation and recognition tasks on OTU\_CEUS become seven-category tasks (not eight-category). When tackling UDA segmentation task, no matter OTU\_2d is served as source or target set, we will remove the ``Normal Ovary'' category from OTU\_2d to make the experiments reasonable. In other words, UDA segmentation task is also a seven-category task. (4) The segmentation and recognition tasks on OTU\_2d are eight-category tasks including ``Normal Ovary'' category. 
\par For almost every images, only one type of tumor appears. Therefore, we transform an seven/eight-category segmentation task into binary lesion area segmentation task (Task1 and Task2) and tumor recognition task (Task3). 
\par On MMOTU image dataset, the pixel-wise semantic annotations and global-wise category annotations are provided by 27 experts of Obstetrics and Gynecology department. Each image is first annotated by one expert and then checked by another one expert, which guarantees the annotating quality. During annotating, experts refer to pathological reports, which makes the annotations accurate and convincing.
\begin{figure}[tb]
\begin{center}
   \includegraphics[width=1.0\linewidth]{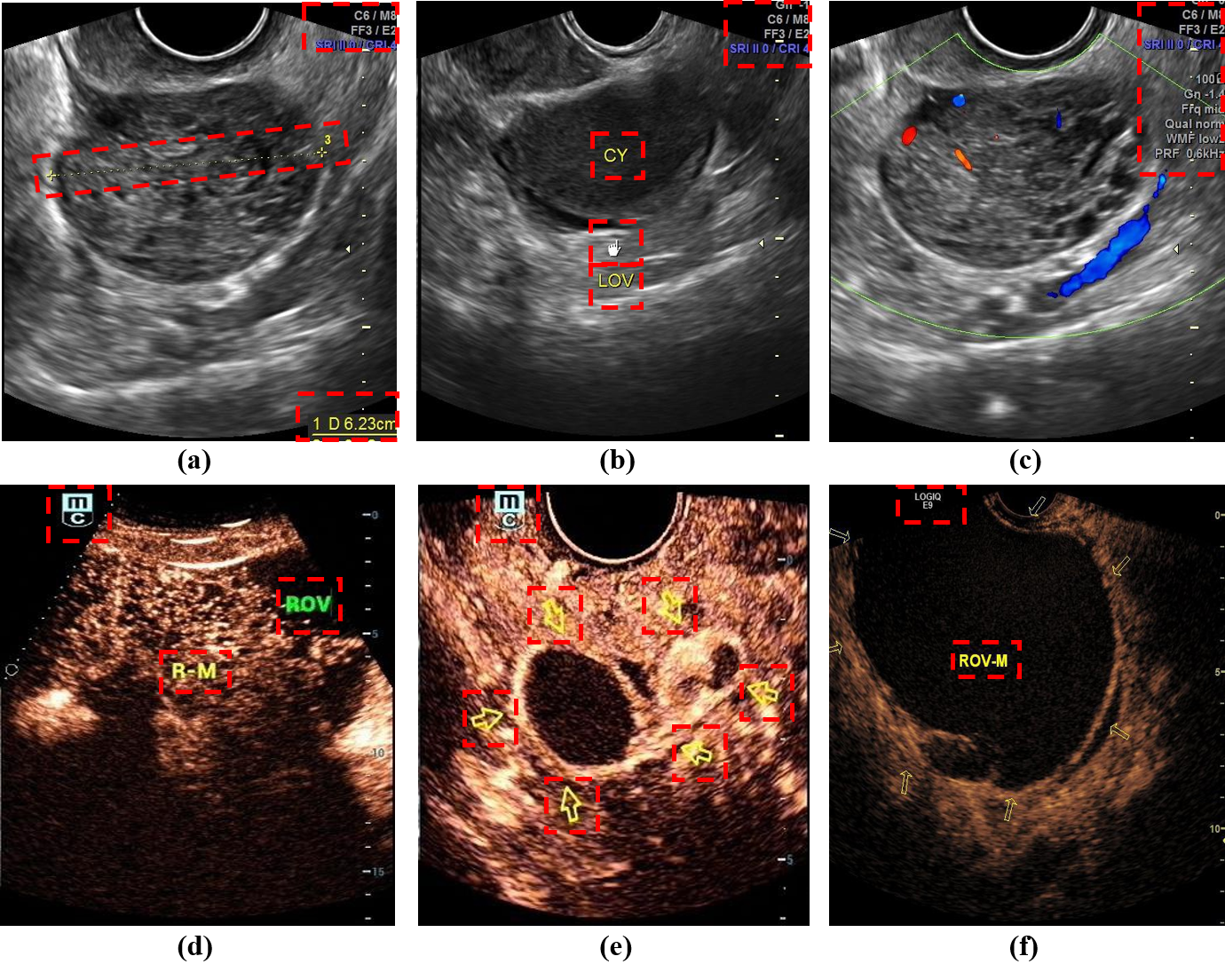}
\end{center}
   \caption{Typical samples in MMOTU image dataset. Images in first and second row are respectively 2d ultrasound and CEUS image samples.}
\label{Fig2}
\end{figure}
\begin{figure}[tb]
\begin{center}
   \includegraphics[width=1.0\linewidth]{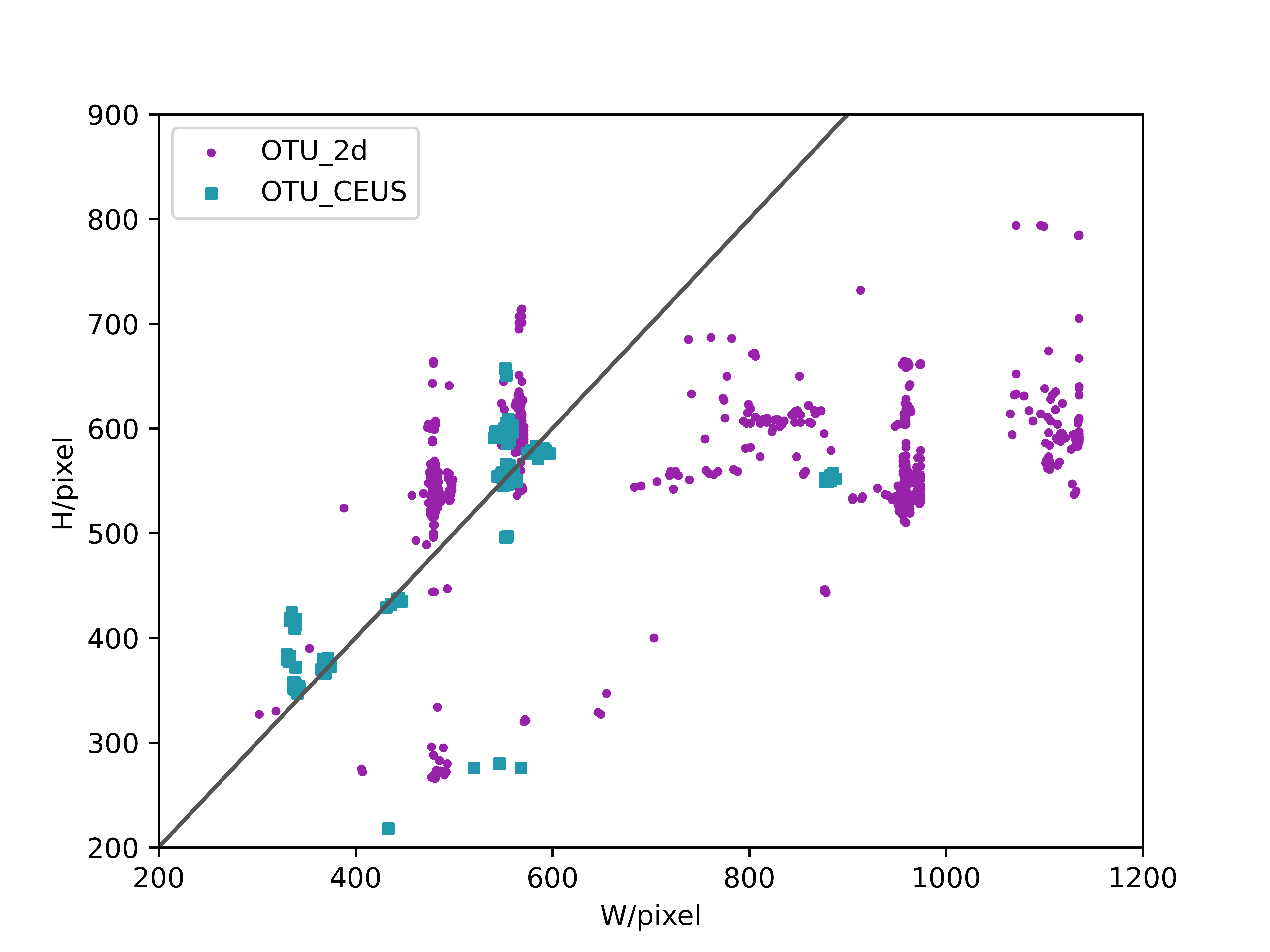}
\end{center}
   \caption{The scatter plot showing the distribution of image scale.}
\label{Fig_1}
\end{figure}
\subsection{Data Analysis}
\par In MMOTU image dataset, all images have been analyzed by experts using specialized software, so some symbols like ``lines'' (Fig.\ref{Fig2}(a)), ``hands'' (Fig.\ref{Fig2}(b)), ``characters'' (Fig.\ref{Fig2}(d) and Fig.\ref{Fig2}(f)) or ``arrows'' (Fig.\ref{Fig2}(e)) are marked by them on images. When we collect the images, most of them contain those symbols. Fig.\ref{Fig2} shows some typical samples. Particularly, OTU\_2d sub-set contains 216 Color Doppler Flow Images (CDFI), where the rest 1253 images are traditional 2d ultrasound images. Images in OTU\_CEUS sub-set are extracted from CEUS sequences. When collecting images, we find that all of them contain private information of patients. According to privacy policy, we manually crop the images to remove the private information and make sure that the published MMOTU image dataset will not contain any private information. For samples from each patient, we provide demographic information including gender, age, weight and doctors' diagnostic description. To protect patients' privacy out of ethics, we think that some information (e.g., age, weight) are not supposed be available. However, to maximally help the community investigate and develop the machine learning models based on MMOTU image dataset, we'd like to make the doctors' diagnostic description available after a research request.
\begin{table}
  \caption{Comparison between MMOTU image dataset and other ovarian ultrasound image dataset. Here, ``vol'' indicates volumes.}
  \centering
  \scalebox{0.7}{
  \begin{tabular}{c c c c c}
  \cmidrule(r){1-5}
  \multirow{2}{*}{Methods} & \multirow{2}{*}{ \shortstack{Data \\ Number}} & \multirow{2}{*}{\shortstack{Category \\ Number}} & \multirow{2}{*}{\shortstack{Data Type \\ (2d/3d/CEUS)}} & \multirow{2}{*}{\shortstack{Task \\ (Cls/Seg/DA-Seg)}} \\ \\
  \cmidrule(r){1-5}
  \cite{DL_OTC} & 988 & 3 & \checkmark/\ding{56}/\ding{56} & \checkmark/\ding{56}/\ding{56} \\
  \cmidrule(r){1-5}
  \cite{Wang_etal} & 412 & 3 & \checkmark/\ding{56}/\ding{56} & \checkmark/\ding{56}/\ding{56} \\
  \cmidrule(r){1-5} 
  \cite{e2e_OTSeg} & 87 & 3 & \checkmark/\ding{56}/\ding{56} & \ding{56}/\checkmark/\ding{56} \\
  \cmidrule(r){1-5}
  \cite{CR-UNet} & 3204 & 3 & \checkmark/\ding{56}/\ding{56} & \ding{56}/\checkmark/\ding{56} \\
  \cmidrule(r){1-5}
  \cite{Narraetal_3d} & 105 (vol) & 3 & \checkmark/\checkmark/\ding{56} & \ding{56}/\checkmark/\ding{56} \\
  \cmidrule(r){1-5}
  \cite{S-Net} & 66 (vol) & 3 & \checkmark/\checkmark/\ding{56} & \ding{56}/\checkmark/\ding{56} \\
  \cmidrule(r){1-5}
  \cite{ovarySeg_MIA} & 307 (vol) & 3 & \checkmark/\checkmark/\ding{56} & \ding{56}/\checkmark/\ding{56} \\
  \cmidrule(r){1-5}
  Ours & 1639 & 8 & \checkmark/\ding{56}/\checkmark & \checkmark/\checkmark/\checkmark \\
  \cmidrule(r){1-5}
  \end{tabular}}
 \label{Tab2}
\end{table}
\par As shown in Fig.\ref{Fig_1}, images in dataset have different scales. In OTU\_2d, the width and height of images respectively range from 302$\sim$1135 and 226$\sim$794 pixels. In OTU\_CEUS, the width and height of images respectively range from 330$\sim$888 and 218$\sim$657 pixels. Before training, images are randomly resized and cropped to 384 $\times$ 384. As mentioned in Sec.\ref{sec:introduction}, we hope to measure the tumor size in single-modality segmentation task. Even though the estimation of the physical size of the tumor will be affected when images' resolutions vary, measurement of tumor size still makes sense when applying MMOTU-trained lesion region segmentation networks on raw images with fixed size.
\par Tab.\ref{Tab2} shows the comparison between MMOTU image dataset and previous ovarian ultrasound image datasets. First, \cite{DL_OTC, Wang_etal} construct ovarian 2d ultrasound datasets to train a model which can categorize the ovarian tumor into Benign, Borderline and Malignant. Comparing with their methods, our dataset contains eight concrete ovarian tumor categories (Fig.\ref{Fig1}). Second, \cite{e2e_OTSeg, CR-UNet, Narraetal_3d, S-Net, ovarySeg_MIA} construct 2d or 3d ultrasound datasets for ovarian structure segmentation. \cite{e2e_OTSeg, CR-UNet} fill their dataset only with 2d ultrasound slices and \cite{Narraetal_3d, S-Net, ovarySeg_MIA} fill their dataset with both 2d ultrasound slices and 3d ultrasound volumes. Strictly speaking, 2d ultrasound slices and 3d ultrasound volumes belong to single-modality data, because 2d ultrasound slices are extracted from a specific 3d ultrasound volume. Comparing with their datasets, our dataset contains 2d ultrasound slices together with CEUS images, which is a typical multi-modality dataset. As a whole, we construct MMOTU image dataset to tackle multi-tasks (Fig.\ref{Fig1}). Moreover, our dataset provides an insight on applying DCNNs for CEUS images.
\section{Proposed Method}
\label{sec:method}
\subsection{Single-Modality Image Segmentation}
\par
\begin{figure}[tb]
\begin{center}
   \includegraphics[width=1.0\linewidth]{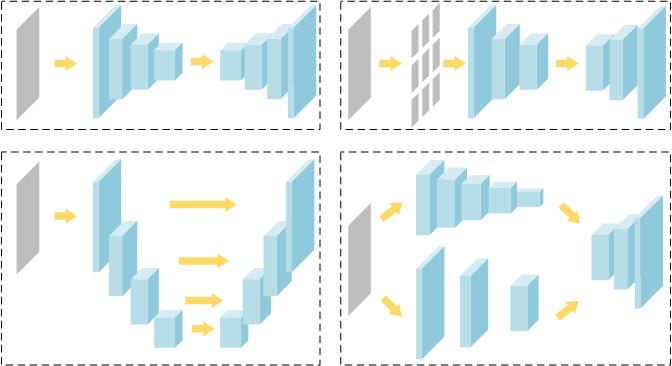}
\end{center}
   \caption{The diagram of recent notable architectures on semantic segmentation. \textbf{top-left:} CNN-based ``Encoder-Decoder'', \textbf{top-right:} Transformer-based ``Encoder-Decoder'', \textbf{bottom-left:} U-shape networks, \textbf{bottom-right:} Spatial-context based two-branch networks.}
\label{Fig3}
\end{figure}

 Based on MMOTU image dataset, we first tackle single-modality semantic segmentation task (Fig.\ref{Fig1} Task1) to provide series of baseline segmentors. Fig.\ref{Fig3} shows the diagrams of four notable architectures in semantic segmentation. For CNN-based ``Encoder-Decoder'', we select two notable segmentors, PSPNet~\cite{PSPNet} and DANet~\cite{DANet}. The former utilizes the spatial-wise contextual information, the latter introduces spatial-wise and channel-wise attention information in segmentors. Transformer-based ``Encoder-Decoder'' is a series of cutting-edge architectures integrating transformers into segmentors to enhance the performance with self-attention mechanism. In this paper, we reimplement SegFormer~\cite{SegFormer}, which is an efficient and lightweight yet powerful semantic segmentation architecture. U-shape networks are most common-used on medical image segmentation. In this paper, we provide two novel segmentors, which are U-Net~\cite{UNet}. and TransUNet~\cite{TransUNet}. As another innovative series of segmentors, spatial-context based two-branch networks use ``wide-and-shallow'' spatial-branch and ``narrow-and-deep'' context-branch to respectively capture low-level high-resolution and high-level context features. Here, we reimplement BiseNetV2~\cite{BiSeNetV2} as representative of this type of segmentors.
\begin{figure*}
\begin{center}
   \includegraphics[width=1.0\linewidth]{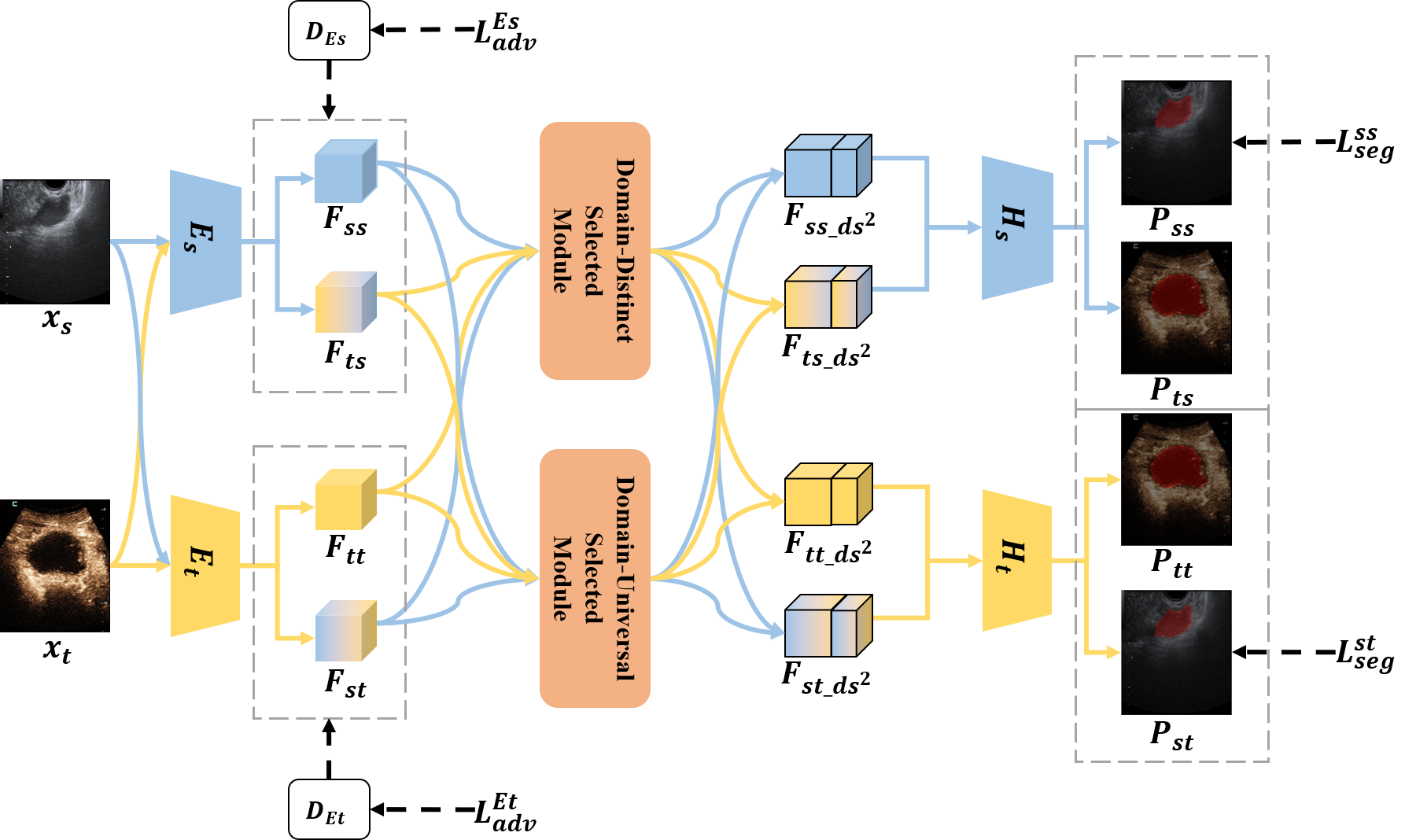}
\end{center}
   \caption{\textbf{The architecture of DS$^2$Net.} Source and target images are first fed into $E_{s}$ and $E_{t}$ and the output feature maps are respectively \{$\bm{F_{ss}}, \bm{F_{st}}$\} and \{$\bm{F_{tt}}, \bm{F_{ts}}$\}. Then, domain distinct selected module (DDSM) and domain universal selected module (DUSM) are applied on \{$\bm{F_{ss}}, \bm{F_{st}}$\} and \{$\bm{F_{tt}}, \bm{F_{ts}}$\}. After processing by DDSM and DUSM, the domain selected features (\{$\bm{F_{ss\_ds^{2}}}, \bm{F_{st\_ds^{2}}}$\}; \{$\bm{F_{tt\_ds^{2}}}, \bm{F_{ts\_ds^{2}}}$\}) pass through $H_{s}$ and $H_{t}$ to generate final predictions. During optimization, we apply symmetric adversarial loss ($\mathcal{L}_{adv}^{E_{s}}, \mathcal{L}_{adv}^{E_{t}}$) and segmentation loss ($\mathcal{L}_{seg}^{ss}, \mathcal{L}_{seg}^{st}$). Specifically, the discriminators \{$D_{E_{s}}, D_{E_{t}}$\} are designed for feature alignment. During inference, we apply ensemble strategy to integrate the two predictions of target images (\{$\bm{P_{tt}}, \bm{P_{ts}}$\}).}
\label{Fig4}
\end{figure*}
\subsection{Unsupervised Domain Adaptation for Semantic Segmentation}
\par Unsupervised domain adaptation (UDA) for semantic segmentation (unsupervised cross-domain semantic segmentation) aims to apply the segmentation network trained from the source domain annotated (labeled) images to the target domain images. It is worth noting that target domain images have no label. Mathematically, given a source domain image dataset $\bm{S} = \{(\bm{x^{i}}, \bm{y^{i}}), \forall i=  1, 2, \dots, N\}$ and a target domain image dataset $\bm{T} = \{\bm{x^{j}}, \forall j=  1, 2, \dots, M\}$, training set $\bm{D_{train}}$ is constructed by $\bm{S}$ and $\bm{T_{train}}$ (a sub-set of $\bm{T}$). Testing set $\bm{D_{test}}$ is constructed by $\bm{T_{test}}$ (a sub-set of $\bm{T}$). Then, $\bm{D_{train}}$ is used to train a segmentation network, which can predict the images in $\bm{D_{test}}$. Here, $\bm{D_{train}} = \bm{S} \cup \bm{T_{train}}$, $\bm{D_{test}} = \bm{T_{test}}$, $\bm{T} = \bm{T_{train}} \cup \bm{T_{test}}$, $\bm{T_{train}} \cap \bm{T_{test}} = \emptyset$, $\bm{S} \cap \bm{T} = \emptyset$. $N$ and $M$ respectively indicate the sample numbers of source and target domain image datasets. Even though target images appear in training process, no label is provided. Therefore, this training process can be regarded as ``unsupervised''.
\par In this paper, exploiting the potential of lesion area segmentation for bidirectional UDA between 2d ultrasound and CEUS images is our main focus. When OTU\_2d is served as source domain image dataset, OTU\_CEUS will be served as target domain image dataset and vice versa. To solve the domain shift problem, we propose a feature alignment based method, DS$^2$Net. The overview of the architecture is shown in Fig.\ref{Fig4}. DS$^2$Net utilizes adversarial learning to alleviate the representation gap between source and target domain images. Based on feature alignment, DS$^2$Net disentangles and fuses 2d ultrasound and CEUS features to tackle multi-modality segmentation task. Essentially, adversarial learning first maps source and target features into same latent space. Then, DDSM and DUSM respectively find domain distinct and universal representations.
\begin{figure*}
\begin{center}
   \includegraphics[width=0.9\linewidth]{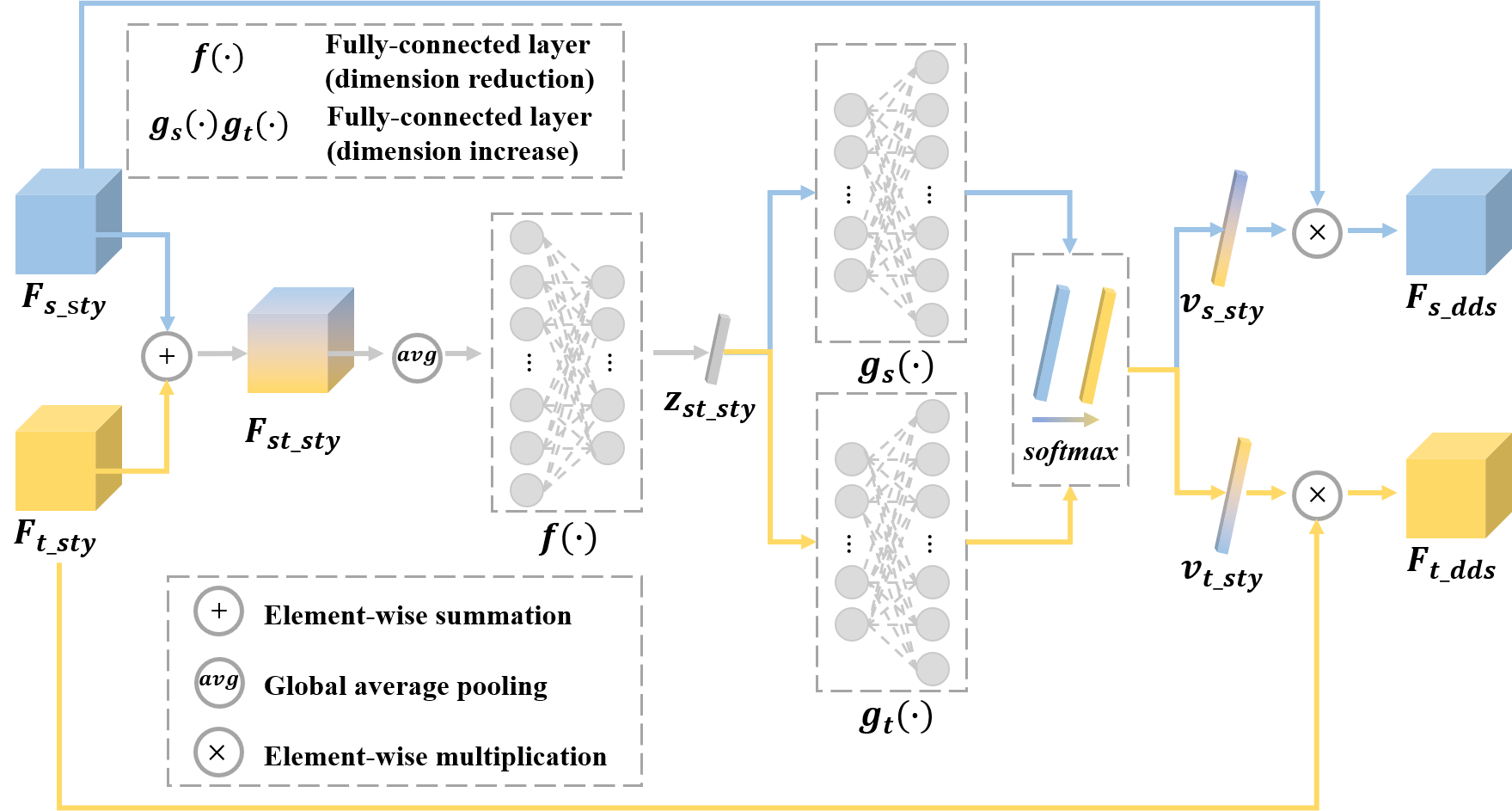}
\end{center}
   \caption{\textbf{The architecture of domain-distinct selected module.} Here, \{$\bm{F_{s\_sty}}, \bm{F_{t\_sty}}$\} are respectively source-style and target-style features of source or target images (\{$\bm{F_{ss}}, \bm{F_{st}}$\} or \{$\bm{F_{tt}}, \bm{F_{ts}}$\}). Two-style features are first integrated to generate a fused prototype ($z_{st\_sty}$). Then, we adopt channel-wise attention to obtain channel-wise weighted vectors (\{$\bm{v_{s\_sty}}, \bm{v_{t\_sty}}$\}) for $\bm{F_{s\_sty}}$ and $\bm{F_{t\_sty}}$. Finally, we multiply \{$\bm{v_{s\_sty}}, \bm{v_{t\_sty}}$\} on $\bm{F_{s\_sty}}$ and $\bm{F_{t\_sty}}$ in channel-wise to obtain domain distinct features (\{$\bm{F_{s\_dds}}, \bm{F_{t\_dds}}$\}).}
\label{Fig5}
\end{figure*}
\subsubsection{Adversarial Learning for Feature Alignment} As shown in Fig.\ref{Fig4}, we respectively design $E_{s}$ and $E_{t}$ for feature extraction of source and target images. Obviously, we expect that both $E_{s}$ and $E_{t}$ have the general representation capability on source and target images. Therefore, we apply adversarial learning on $E_{s}$ and $E_{t}$ to alleviate the representation gap caused by domain shift. This process is formulated from Eq.\ref{Eq1} to Eq.\ref{Eq4}.
\begin{equation}
\begin{split}
   \mathcal{L}_{adv}^{E_{s}}(E_{s}, D_{E_{s}}) &= \mathbb{E}_{x^{s} \sim X^{s}}[log(D_{E_{s}}(\bm{F_{ss}}))]  \\
                  &+ \mathbb{E}_{x^{t} \sim X^{t}}[log(1 - D_{E_{s}}(\bm{F_{ts}}))]
   \end{split}
\label{Eq1}
\end{equation}
\begin{equation}
\bm{F_{ss}} = E_{s}(\bm{x_{s}}),\quad \bm{F_{ts}} = E_{s}(\bm{x_{t}})
\label{Eq2}
\end{equation}
\begin{equation}
\begin{split}
   \mathcal{L}_{adv}^{E_{t}}(E_{t}, D_{E_{t}}) &= \mathbb{E}_{x^{t} \sim X^{t}}[log(D_{E_{t}}(\bm{F_{tt}}))] \\ 
   &+ \mathbb{E}_{x^{s} \sim X^{s}}[log(1 - D_{E_{t}}(\bm{F_{st}}))] 
\end{split}
\label{Eq3}
\end{equation}
\begin{equation}
\bm{F_{tt}} = E_{t}(\bm{x_{t}}),\quad \bm{F_{st}} = E_{t}(\bm{x_{s}})
\label{Eq4}
\end{equation}
\par Here, $\bm{x_{s}}$ and $\bm{x_{t}}$ respectively denote source and target images. $D_{E_{s}}$ and $D_{E_{t}}$ are two discriminators. $\bm{F_{ss}}, \bm{F_{ts}}, \bm{F_{tt}}, \bm{F_{st}}$ are respectively denoted as source-style source feature, source-style target feature, target-style target feature and target-style source feature. We adopt min-max criterion to optimize $E_{s}$ and $E_{t}$, formulated as $min_{E_{s}}max_{D_{E_{s}}}\mathcal{L}_{adv}^{E_{s}}(E_{s}, D_{E_{s}})$ and $min_{E_{t}}max_{D_{E_{t}}}\mathcal{L}_{adv}^{E_{t}}(E_{t}, D_{E_{t}})$. With this feature alignment on $E_{s}$ and $E_{t}$, two encoders have better ability to represent images from two domains. Furthermore, two encoders can provide one more feature map for image from each domain, which enriches the representations. 
\subsubsection{Domain Distinct \& Domain Universal Selected Modules} As shown in Eq.\ref{Eq2} and Eq.\ref{Eq4}, we design two encoders to extract source-style and target-style features for source and target images. Based on the observation that directly applying a source-trained segmentation model on target images will not cause dramatic performance decrease, we make the following assumptions. (1) Between 2d ultrasound and CEUS images, there may exist some easy-extracted universal features, which have positive influence on segmenting images from both two domains. (2) Since there still exists some performance gap when simply evaluating target images on source-trained models, we believe that domain distinct features also exist. (3) Distinct and universal features can be decoupled from source-style and target-style features. Instructed by the above assumptions, we design Domain Distinct Selected Modules (DDSM) and Domain Universal Selected Modules (DUSM).
\par \textbf{DDSM:} As shown in Fig.\ref{Fig4}, source and target images are both represented as source-style and target-style features, which respectively are \{$\bm{F_{ss}}, \bm{F_{st}}$\} ($\mathbb{R}^{C \times H \times W}$) and \{$\bm{F_{tt}}, \bm{F_{ts}}$\} ($\mathbb{R}^{C \times H \times W}$). After applying adversarial learning, $E_{s}$ and $E_{t}$ can better represent two domain images as source-style and target-style features. To further extract purer feature specifically for each domain, we propose DDSM.
\par As shown in Fig.\ref{Fig5} and Eq.\ref{Eq5}, we first fuse source-style and target-style features. Then, we apply global average pooling and a dimension-reduction fully-connected layer ($f(\cdot)$) to generate a global-wise fused prototype ($\bm{z_{st\_sty}} \in \mathbb{R}^{C' \times 1 \times 1}$).
\begin{figure*}
\begin{center}
   \includegraphics[width=0.9\linewidth]{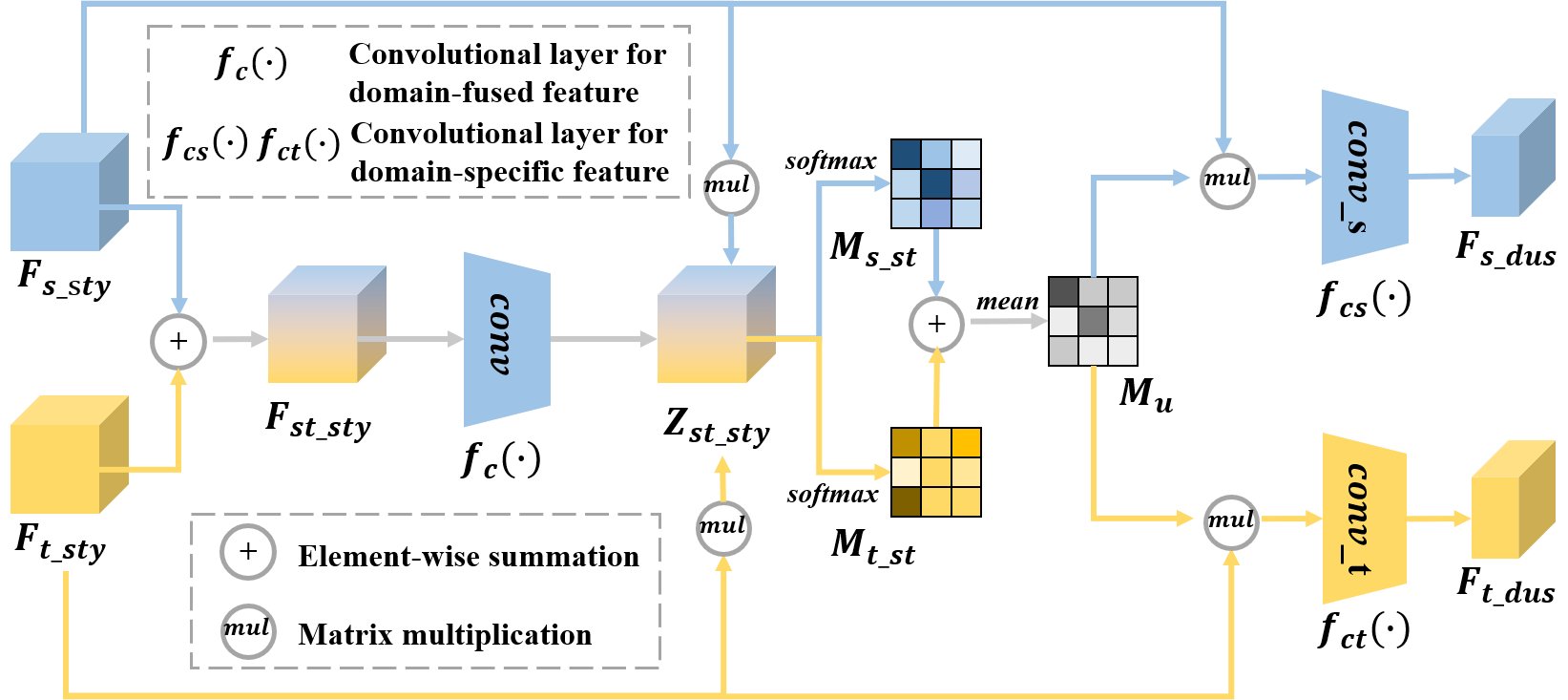}
\end{center}
   \caption{\textbf{The architecture of domain-universal selected module.} Two-style features are first integrated to generate a fused feature ($Z_{st\_sty}$). Then, we adopt matrix multiplication operation to obtain attention masks (\{$\bm{M_{s\_sty}}, \bm{M_{t\_sty}}$\}) for $\bm{F_{s\_sty}}$ and $\bm{F_{t\_sty}}$. Next, we average $\bm{M_{s\_sty}}$ and $\bm{M_{t\_sty}}$ for a universal attention mask ($\bm{M_{u}}$). Finally, we serve $\bm{M_{u}}$ as universal guidance on $\bm{F_{s\_sty}}$ and $\bm{F_{t\_sty}}$ to obtain domain universal features (\{$\bm{F_{s\_dus}}, \bm{F_{t\_dus}}$\}). }
\label{Fig6}
\end{figure*}
\begin{small}
\begin{equation}
\begin{split}
\bm{z_{st\_sty}} &= f(\frac{1}{HW}\sum_{h=1}^{H}\sum_{w=1}^{W}(\bm{F_{st\_sty}}(h, w))) \\ &= f(\frac{1}{HW}\sum_{h=1}^{H}\sum_{w=1}^{W}(\bm{F_{s\_sty}}(h, w) + \bm{F_{t\_sty}}(h, w)))
\end{split}
\label{Eq5}
\end{equation}
\end{small}
\par Then, we serve $\bm{z_{st\_sty}}$ as guidance to create channel-wise weighted vectors ($\bm{v_{s\_sty}}$ and $\bm{v_{t\_sty}} \in \mathbb{R}^{C \times 1 \times 1}$) using $softmax$ operation shown in Eq.\ref{Eq6} and Eq.\ref{Eq7}. $g_{s}(\cdot)$ and $g_{t}(\cdot)$ are dimension-increase fully-connected layers.

\begin{equation}
\bm{v_{s\_sty}} = \frac{e^{g_{s}(\bm{z_{st\_sty}})}}{e^{g_{s}(\bm{z_{st\_sty}})} + e^{g_{t}(\bm{z_{st\_sty}})}}
\label{Eq6}
\end{equation}
\begin{equation}
\bm{v_{t\_sty}} = \frac{e^{g_{t}(\bm{z_{st\_sty}})}}{e^{g_{s}(\bm{z_{st\_sty}})} + e^{g_{t}(\bm{z_{st\_sty}})}}
\label{Eq7}
\end{equation}
\par Finally, the domain distinct selected features ($\bm{F_{s\_dds}}$ and $\bm{F_{t\_dds}} \in \mathbb{R}^{C \times H \times W}$) are calculated through channel-wise multiplication shown in Eq.\ref{Eq8}. 
\begin{equation}
\bm{F_{s\_dds}} = \bm{F_{s\_sty}} \times \bm{v_{s\_sty}}, \quad \bm{F_{t\_dds}} = \bm{F_{t\_sty}} \times \bm{v_{t\_sty}}
\label{Eq8}
\end{equation}
\par From Eq.\ref{Eq6} to Eq.\ref{Eq7}, we find that $\bm{v_{s\_sty}}+\bm{v_{t\_sty}}=\bm{1}$, which means we force the source-style and target-style features to have complementary channel combinations. With this design, ``source-style features are more source-style while target-style features are more target-style''.
\par \textbf{DUSM:} With DDSM, we obtain purer source-style and target-style features, but it may overlook the universal features benefiting both source and target images. Therefore, we further design DUSM to extract universal features of $\bm{F_{s\_sty}}$ and $\bm{F_{t\_sty}}$. As shown in Fig.\ref{Fig6}, we first fuse source-style and target-style features similar to DDSM (Eq.\ref{Eq9}).

\begin{small}
\begin{equation}
\begin{split}
\bm{Z_{st\_sty}} &= f_{c}(\frac{1}{HW}\sum_{h=1}^{H}\sum_{w=1}^{W}(\bm{F_{st\_sty}}(h, w))) \\ &= f_{c}(\frac{1}{HW}\sum_{h=1}^{H}\sum_{w=1}^{W}(\bm{F_{s\_sty}}(h, w) + \bm{F_{t\_sty}}(h, w)))
\end{split}
\label{Eq9}
\end{equation}
\end{small}
\par Here, $\bm{Z_{st\_sty}}$ ($\mathbb{R}^{C \times H \times W}$) represent domain fused feature. We then perform matrix multiplication between $\bm{Z_{st\_sty}}$ and \{$\bm{F_{s\_sty}}, \bm{F_{t\_sty}}$\}. The output attention masks (\{$\bm{M_{s\_st}}, \bm{M_{t\_st}}$\} $\in$ $\mathbb{R}^{C \times C}$) are averaged for universal attention mask ($\bm{M_{u}}$ $\in$  $\mathbb{R}^{C \times C}$). This process is shown in Eq.\ref{Eq10} $\sim$ Eq.\ref{Eq12}.
\begin{equation}
M_{s\_st}(j, i) = \frac{e^{F_{s\_sty}(i) \cdot Z_{st\_sty}(j)}}{\sum_{i=1}^{C}(e^{F_{s\_sty}(i) \cdot Z_{st\_sty}(j)})}
\label{Eq10}
\end{equation}
\begin{equation}
M_{t\_st}(j, i) = \frac{e^{F_{t\_sty}(i) \cdot Z_{st\_sty}(j)}}{\sum_{i=1}^{C}(e^{F_{t\_sty}(i) \cdot Z_{st\_sty}(j)})}
\label{Eq11}
\end{equation}
\begin{equation}
M_{u}(j, i) = \frac{M_{s\_st}(j, i) + M_{t\_st}(j, i)}{2}
\label{Eq12}
\end{equation}
where $\bm{F_{s\_sty}}$, $\bm{F_{t\_sty}}$ and $\bm{Z_{st\_sty}}$ are reshaped to $\mathbb{R}^{C \times (HW)}$. ``$(j, i)$'' indicates the impact measurement between $i^{th}$ channel of \{$\bm{F_{s\_sty}}, \bm{F_{t\_sty}}$\} and $j^{th}$ channel of $\bm{Z_{st\_sty}}$.
\par Finally, we use $\bm{M_{u}}$ to extract the universal feature from $\bm{F_{s\_sty}}$ and $\bm{F_{t\_sty}}$, which are $\bm{F_{s\_dus}}$ and $\bm{F_{t\_dus}}$ ($\mathbb{R}^{C'' \times H \times W}$). This formulation is shown in Eq.\ref{Eq13}.

\begin{small}
\begin{equation}
\bm{F_{s\_dus}} = f_{cs}(\bm{M_{u}} * \bm{F_{s\_sty}}), \quad \bm{F_{t\_dus}} = f_{ct}(\bm{M_{u}} * \bm{F_{t\_sty}})
\label{Eq13}
\end{equation}
\end{small}
\par In summary, we extract the universal feature by finding $\bm{M_{u}}$, which measures the joint impact between source/target-style features and domain fused feature. The larger element in $\bm{M_{u}}$ indicates that one channel of fused domain features has high impact both on same channel of source-style and target-style features. Essentially, we utilize a domain universal attention mask to select the universal features in channel-wise manner.
\par \textbf{Analysis between DDSM and DUSM:} Generally, two modules both focus on improving the feature-level understanding for target domain images. The main differences between DDSM and DUSM can be listed in the following aspects. (1) Motivation differences. Motivated by decoupling and extracting domain-distinct features, we design DDSM. Motivated by selecting and extracting domain-universal features, we design DUSM. (2) Designing principle differences. As shown in Fig.\ref{Fig5}, DDSM decouples fused features ($\bm{F_{st\_sty}}$) by extracting opposing and complementary channel-wise attention (channel-wise weighted vectors) for two-style features. As shown in Fig.\ref{Fig6}, DDUM extracts universal features by finding the channel-wise universal attention mask ($\bm{M_{u}}$). (3) Feature extraction differences. DDSM utilizes global-level features to decouple distinct features while DUSM utilizes semantic-level features to finding universal features.
\subsubsection{Optimization} In this paper, we apply segmentation loss and adversarial loss to jointly optimize the DS$^2$Net. As for segmentation loss, We apply cross-entropy loss, formulated in Eq.\ref{Eq14}.

\begin{small}
\begin{equation}
\begin{split}
   \mathcal{L}_{seg}(E_{s}, H_{s}, E_{t}, H_{t}) &= \mathcal{L}_{seg}^{ss}(E_{s}, H_{s}) + \mathcal{L}_{seg}^{st}(E_{t}, H_{t}) \\ &= \mathbb{E}_{x^{s} \sim X^{s}}[-y_{s}log(H_{s}(\bm{F_{ss\_ds^{2}}}))] \\
                  &+ \mathbb{E}_{x^{s} \sim X^{s}}[-y_{s}log(H_{t}(\bm{F_{st\_ds^{2}}}))]
   \end{split}
\label{Eq14}
\end{equation}
\end{small}
where $H_{s}$ and $H_{t}$ are two decoder-heads respectively designed for source-style and target-style features. $\bm{F_{ss\_ds^{2}}} = cat(\bm{F_{ss\_dds}}, \bm{F_{ss\_dus}}); \quad  \bm{F_{st\_ds^{2}}} = cat(\bm{F_{st\_dds}}, \bm{F_{st\_dus}})$. ``cat'' denotes concatenation. As shown in Eq.\ref{Eq8}, Eq.\ref{Eq13}, $\{\bm{F_{ss\_ds^{2}}}, \bm{F_{st\_ds^{2}}}\}$ $\in \mathbb{R}^{(C+C'') \times H \times W}$.
\par To optimize the whole architecture, we apply combined loss ($\mathcal{L})$ shown in Eq.\ref{Eq15}. $\{\lambda_{adv}^{E_{t}}, \lambda_{adv}^{E_{t}}\}$ are trade-off hyper-parameters adjusting the impact of each loss item.
\begin{equation}
\begin{split}
    \mathcal{L} &= \mathcal{L}_{seg}(E_{s}, H_{s}, E_{t}, H_{t}) \\ &+ \lambda_{adv}^{E_{s}} \mathcal{L}_{adv}^{E_{s}}(E_{s}, D_{E_{s}}) + \lambda_{adv}^{E_{t}} \mathcal{L}_{adv}^{E_{t}}(E_{t}, D_{E_{t}})
\end{split}
    \label{Eq15}
\end{equation}
\par From the perspective of backward optimization process, we find the following points. (1) Through adversarial learning, $E_{s}$ and $E_{t}$ have the basic representation ability on source and target images avoiding representation distortion of target images. (2) Based on adversarial learning and DDSM/DUSM, the segmentation loss ($\mathcal{L}_{seg}^{ss}(E_{s}, H_{s})$ and $\mathcal{L}_{seg}^{st}(E_{t}, H_{t})$) can force \{$E_{s}, H_{s}$\} and \{$E_{t}, H_{t}$\} to have specific representation trend (e.g. \{$E_{t}, H_{t}$\} tends to represent source and target images into target style). Optimized like this, the architecture becomes more friendly to target images.
\subsubsection{Inference} As shown in Eq.\ref{Eq2} and Eq.\ref{Eq4}, target image ($\bm{x_{t}}$) first passes through $E_{s}$ and $E_{t}$ to generate $\bm{F_{ts}}$ and $\bm{F_{tt}}$. Then $\bm{F_{ts}}$ and $\bm{F_{tt}}$ are respectively fed into DDSM and DUSM to generate domain distinct features (\{$\bm{F_{ts\_dds}}, \bm{F_{tt\_dds}}$\}) and domain universal features (\{$\bm{F_{ts\_dus}}, \bm{F_{tt\_dus}}$\}). Next, different style features are concatenated together, which can be expressed as $\bm{F_{ts\_ds^{2}}} = cat(\bm{F_{ts\_dds}}, \bm{F_{ts\_dus}})$ and $\bm{F_{tt\_ds^{2}}} = cat(\bm{F_{tt\_dds}}, \bm{F_{tt\_dus}})$. Finally, we use two decoder-heads to map these features into two predictions (\{$\bm{P_{ts}}, \bm{P_{tt}}$\}) and integrate them for final prediction ($\bm{P_{t}}$), shown in Eq.\ref{Eq16}.
\begin{equation}
\bm{P_{t}} = \frac{\bm{P_{ts}} + \bm{P_{tt}}}{2} = \frac{H_{s}(\bm{F_{ts\_ds^{2}}}) + H_{t}(\bm{F_{tt\_ds^{2}}})}{2}
\label{Eq16}
\end{equation}
\par Here, $\bm{P_{tt}}$ and $\bm{P_{ts}}$ are $softmax$ predictions (posterior probabilities). Inspired by ensemble learning, we apply soft voting to integrate two predictions. The predictive segmentation mask can be calculated by $\mathop{\arg\max}(\cdot)$ operation.
\begin{table}
  \caption{Implementation details of single-modality semantic segmentation. ``$\dag$'' means loading ImageNet-pretrained weights~\cite{ImageNet}. ``Res50'', ``mit\_b5'' ``ViT'' respectively denote ResNet-50~\cite{ResNet}, Mix Transformer~\cite{SegFormer} and Vision Transformer~\cite{ViT}.}
  \centering
  \setlength{\tabcolsep}{1mm}{
  \begin{tabular}{c c c c c c}
  \cmidrule(r){1-6}
  \multirow{2}{*}{Methods} & \multirow{2}{*}{Optimizer} & \multirow{2}{*}{\shortstack{Initial \\ LR}} & \multirow{2}{*}{\shortstack{LR \\ Decay}} & \multirow{2}{*}{Backbone} & \multirow{2}{*}{Iterations} \\ \\
  \cmidrule(r){1-6}
  PSPNet & SGD & 0.01 & poly & Res50$^{\dag}$ & 80k \\
  \cmidrule(r){1-6}
  DANet & SGD & 0.01 & poly & Res50$^{\dag}$ & 80k \\
  \cmidrule(r){1-6}
  SegFormer & Adam & 0.00006 & poly & mit\_b5$^{\dag}$ & 80k \\
  \cmidrule(r){1-6}
  U-Net & SGD & 0.01 & poly & U-Net$^{\dag}$ & 80k \\
  \cmidrule(r){1-6}
  TransUNet & SGD & 0.01 & poly & Res50+ViT$^{\dag}$ & 80k \\
  \cmidrule(r){1-6}
  BiseNetV2 & SGD & 0.05 & poly & BiseNetV2$^{\dag}$ & 80k \\
  \cmidrule(r){1-6}
  \end{tabular}}
 \label{Tab3}
\end{table}
\section{Experiments and Analysis}
\label{sec:experiments}
\subsection{Single-Modality Semantic Segmentation}
\subsubsection{Implementation Details} As mentioned in Sec.\ref{sec:method} and Fig.\ref{Fig3}, we reimplement the six baseline segmentors on MMOTU image dataset. Before training, input images are randomly resized and cropped to 384 $\times$ 384. A series of data augmentation methods, such as ``RandomFlip'' are applied. Towards the four types of segmentors, the implementation details are shown in Tab.\ref{Tab3}. Here, when optimizer is SGD, the momentum is set as 0.9 and the weight decay value is 0.0005. When optimizer is Adam, the weight decay value is 0.01. All segmentors adopt ``poly'' to adjust learning rate and the power is set to 0.9.
\par As shown in Tab.\ref{Tab3}, SegFormer is optimized with Adam while other segmentation networks are optimized with SGD. Moreover, different networks use different learning rates. To reimplement baseline segmentors on single-modality semantic segmentation task for our proposed MMOTU image dataset, we mainly follow these methods' original training implementation details, which leads to different configurations in Tab.\ref{Tab3}. Technically, we select six baseline segmentation networks according to four types of architectures (Fig.\ref{Fig3}). Among the six networks, SegFormer belongs to Transformer-based network while PSPNet, DANet, U-Net and BiseNetV2 belong to CNN-based networks. Transformer and CNN have large structural discrepancy, so we select suitable optimizer and learning rate based on previous knowledge and experimental experience to ensure that each network can boost its performance. In addition, TransUNet is a CNN-Transformer integrated network. Since large number of trainable parameters are from CNN, we select same configuration as CNN-based networks.
\begin{table}
  \caption{Single-modality segmentation results on MMOTU image dataset. We adopt $IoU$ (Intersection over Union) as metric.}
  \centering
  \begin{tabular}{c c c c c}
  \cmidrule(r){1-5}
  \multirow{2}{*}{Methods}    &\multicolumn{2}{c}{OTU\_2d}        &\multicolumn{2}{c}{OTU\_CEUS}   \\
  \cmidrule(r){2-5}
                         &$IoU (\%)$   &$mIoU (\%)$
                         &$IoU (\%)$   &$mIoU (\%)$      \\
  \cmidrule(r){1-5}
  PSPNet & 82.24  & 89.66 & 71.35  & 81.64 \\
  \cmidrule(r){1-5}
  DANet    &\textbf{82.58}  & \textbf{89.97}  & 70.69  & 81.87 \\
  \cmidrule(r){1-5}
  SegFormer &82.46   &89.88  &\textbf{73.03}  &\textbf{83.00}  \\
  \cmidrule(r){1-5}
  U-Net &79.91  &86.80  &69.18 &80.04 \\
  \cmidrule(r){1-5}
  TransUNet &81.31  &89.01  &70.15  &80.82 \\
  \cmidrule(r){1-5}
  BiseNetV2 &79.37  &86.13  &70.25  &80.98 \\
  \cmidrule(r){1-5}
  \end{tabular}
 \label{Tab4}
\end{table}
\subsubsection{Experimental Results} The single-modality segmentation results are shown in Tab.\ref{Tab4}. The $IoU$ value indicates the intersection of union between predicted and annotated lesion area. The $mIoU$ value indicates the average $IoU$ value of lesion area and background. Tab.\ref{Tab4} shows that U-Net and BiseNetV2 perform poorer compared to other methods. DANet and SegFormer respectively outperform other methods on OTU\_2d and OTU\_CEUS. Particularly on OTU\_CEUS, with less training samples, SegFormer shows much stronger performance. From Tab.\ref{Tab4}, we can find that PSPNet and SegFormer perform more balanced on two sub-sets. Therefore, we select them as baseline segmentors to construct DS$^2$Net for the following cross-domain semantic segmentation task.
\begin{table}
  \caption{Training details of cross-domain semantic segmentation. DS$^2$Net\_P and DS$^2$Net\_T respectively denote the DS$^2$Net following PSPNet and SegFormer.}
  \centering
  \resizebox{\linewidth}{!}{
  \begin{tabular}{c c c c c c c}
  \cmidrule(r){1-7}
  \multirow{3}{*}{Methods}    &\multicolumn{2}{c}{Encoders/DDSM/DUSM}
  &\multicolumn{2}{c}{Decoder-Heads}
  &\multicolumn{2}{c}{Discriminator} \\
  \cmidrule(r){2-7}
                         &\multirow{2}{*}{optimizer}   &\multirow{2}{*}{\shortstack{initial \\ (LR)}}
                         &\multirow{2}{*}{optimizer}   &\multirow{2}{*}{\shortstack{initial \\ (LR)}}
                         &\multirow{2}{*}{optimizer}   &\multirow{2}{*}{\shortstack{initial \\ (LR)}} \\ \\
  \cmidrule(r){1-7}
  DS$^2$Net\_P  &SGD  &0.01  &SGD  &0.02 &Adam &0.00025\\ 
  \cmidrule(r){1-7}
  DS$^2$Net\_T  &Adam  &0.00006  &Adam  &0.00001 &Adam 
  &0.00001 \\
  \cmidrule(r){1-7}
  \end{tabular}}
 \label{Tab5}
\end{table}
\subsection{Cross-Domain Semantic Segmentation}
\subsubsection{Implementation Details} As shown in Fig.\ref{Fig4}, DS$^2$Net is a symmetric ``Encoder-Decoder'' architecture. Since PSPNet and SegFormer are notable ``Encoder-Decoder'', we follow them to design DS$^2$Net. Specifically, DS$^2$Net\_P and DS$^2$Net\_T respectively inherit the encoder and decoder of PSPNet and SegFormer. DS$^2$Net\_P adopts ResNet-50 as encoders ($E_{s}$, $E_{t}$) and ``PSPHead'' as decoder-heads ($H_{s}$, $H_{t}$). DS$^2$Net\_T adopts ``Hierarchical Transformer'' as encoders and ``All-MLP'' as decoder-heads. As shown in Tab.\ref{Tab5}, DS$^2$Net\_P and DS$^2$Net\_T use different implementation configurations mainly because of the large structural discrepancy between Transformer and CNN. For two discriminators (\{$D_{E_{s}}, D_{E_{t}}$\}), we follow the common-applied configuration of PatchGAN~\cite{pix2pix}. \{$D_{E_{s}}, D_{E_{t}}$\} consist of 4 convolutional layers with kernels as size of 4 $\times$ 4. The stride of the first two layers is 2 while the the last two layers set the stride as 1. The output channels of each layer are \{64, 128, 256, 1\}. The training details are shown in Tab.\ref{Tab5}. Additionally, we select binary cross entropy loss as adversarial loss function and the training factors ($\{\lambda_{adv}^{E_{t}}, \lambda_{adv}^{E_{t}}\}$) are set as 0.005.
\begin{table}
  \caption{Comparison between single-modality segmentation and no domain adaptation (w/o DA) segmentation. The left and right of $\rightarrow$ indicate source and target dataset respectively. Results of PSPNet and SegFormer are obtained by training and evaluating on single-modality target dataset.}
  \centering
  \resizebox{\linewidth}{!}{
  \begin{tabular}{c c c c c}
  \cmidrule(r){1-5}
  \multirow{2}{*}{Methods}    &\multicolumn{2}{c}{OTU\_CEUS $\rightarrow$ OTU\_2d}        &\multicolumn{2}{c}{OTU\_2d $\rightarrow$ OTU\_CEUS}   \\
  \cmidrule(r){2-5}
                         &$IoU (\%)$   &$mIoU (\%)$
                         &$IoU (\%)$   &$mIoU (\%)$      \\
  \cmidrule(r){1-5}
  PSPNet  & 82.24  & 89.66 & 71.35 & 81.64 \\
  \cmidrule(r){1-5}
  PSPNet (w/o DA)  &  44.17  & 66.95 & 50.22  & 67.60 \\
  \cmidrule(r){1-5}
  SegFormer &82.46  &89.88  &73.03  &83.00  \\
  \cmidrule(r){1-5}
  SegFormer (w/o DA) &56.28  &74.04 &61.11 &75.45  \\
  \cmidrule(r){1-5}
  \end{tabular}}
 \label{Tab6}
\end{table}
\subsubsection{No Domain Adaptation Experiments} Before evaluating the domain adaptation performance of DS$^2$Net on MMOTU image dataset, we first conduct no domain adaptation experiment. In this experiment, PSPNet and SegFormer are trained by source annotated samples and evaluated on target samples using the configurations in Tab.\ref{Tab3}. The results are shown in Tab.\ref{Tab6}. Previous methods~\cite{SIFA, DSFN} apply no domain adaptation experiment between CT and MR images. The segmentation results always dramatically decrease (40$\sim$60\% lower than single-modality training). Compared to previous results, our results on domain adaptation do not decrease by large margin, E.g., when SegFormer is trained with OTU\_2d and evaluated on OTU\_CEUS, the $IoU$ value decreases by 11.92\% (73.03\% $\sim$ 61.11\%) comparing with target-trained SegFormer. It worth noting that the results training with OTU\_2d and evaluating on OTU\_CEUS are more valuable, because OTU\_2d has more training samples than OTU\_CEUS (Tab.\ref{Tab1}). No domain adaptation experiments results inspire us to explore the potential of feature alignment and distinct/universal feature extraction.
\begin{table}
  \caption{5-Fold cross-validation tests on UDA cross-modality ovarian lesion area semantic segmentation. Results are reported as ``mean $\pm$ standard deviation''.}
  \centering
  \resizebox{\linewidth}{!}{
  \begin{tabular}{c c c c c}
  \cmidrule(r){1-5}
  \multirow{2}{*}{Methods}    &\multicolumn{2}{c}{OTU\_CEUS $\rightarrow$ OTU\_2d}        &\multicolumn{2}{c}{ OTU\_2d $\rightarrow$ OTU\_CEUS}   \\
  \cmidrule(r){2-5}
                         &$IoU (\%)$   &$mIoU (\%)$
                         &$IoU (\%)$   &$mIoU (\%)$      \\
  \cmidrule(r){1-5}
  AdapSegNet\_P  & 47.50$\pm$2.73 & 66.15$\pm$2.51 & 55.08$\pm$2.12 & 70.24$\pm$1.83 \\
  \cmidrule(r){1-5}
  EGUDA\_P & 48.91$\pm$1.85 & 67.94$\pm$1.89 & 56.76$\pm$1.59 & 72.19$\pm$1.40 \\
  \cmidrule(r){1-5}
  DS$^2$Net\_P (ours) & 52.73$\pm$1.51 & 71.87$\pm$1.35 & 60.21$\pm$1.26 & 74.38$\pm$1.01 \\
  \cmidrule(r){1-5}
  \cmidrule(r){1-5}
  AdapSegNet\_T  & 62.37$\pm$3.29 & 76.06$\pm$2.46 & 63.17$\pm$2.27 & 76.17$\pm$2.11 \\
  \cmidrule(r){1-5}
  EGUDA\_T  & 58.06$\pm$2.62 & 73.12$\pm$1.99 & 59.72$\pm$1.96 & 74.78$\pm$2.80 \\
  \cmidrule(r){1-5}
   DDFSeg  & 55.92$\pm$5.38 & 72.85$\pm$4.39 & 60.94$\pm$4.13 & 74.59$\pm$3.53 \\
  \cmidrule(r){1-5}
  DAFormer  & 62.76$\pm$2.78 & 76.52$\pm$2.14 & 58.52$\pm$2.41 & 73.97$\pm$2.61 \\
  \cmidrule(r){1-5}
  DCLA  & 61.53$\pm$4.79 & 75.31$\pm$4.67 & 60.20$\pm$5.16 & 74.88$\pm$4.74 \\
  \cmidrule(r){1-5}
  EDRL  & 63.05$\pm$6.13 & 76.63$\pm$5.45 & 57.41$\pm$4.86 & 72.95$\pm$4.07 \\
  \cmidrule(r){1-5}
  DS$^2$Net\_T (ours) & \textbf{64.67$\pm$2.34} & \textbf{78.12$\pm$1.97} & \textbf{68.73$\pm$1.67} & \textbf{79.59$\pm$1.50} \\
  \cmidrule(r){1-5}
  \end{tabular}}
 \label{Tab7}
\end{table}
\subsubsection{Cross-Domain Semantic Segmentation Experiments} To prove the effectiveness of DS$^2$Net, we re implement two notable feature-alignment based methods, which are AdapSegNet~\cite{AdapSegNet} and EGUDA~\cite{EGUDA} on MMOTU image dataset. For fairly comparison, minor modifications are applied. For AdapSegNet and EGUDA, we respectively replace the original shared segmentation network and segmentor with PSPNet or SegFormer. For the discriminator, we adopt the same PatchGAN as \{$D_{E_{s}}, D_{E_{t}}$\} of DS$^2$Net. We further reimplement 4 recent notable methods (DDFSeg~\cite{DDFSeg}, DAFormer~\cite{DAFormer}, EDRL~\cite{EDRL} and DCLA~\cite{DCLA}) following their original model configurations. Specifically, DDFSeg, EDRL and DCLA are typical image alignment methods with feature disentanglement mechanism. DAFormer is a novel self-training based method. In addition, these 4 methods are not feature-alignment methods with typical ``Encoder-Decoder'' architecture.
\par To rigorously compare DS$^2$Net with previous methods, we conduct 5-Fold cross-validation tests on cross-domain semantic segmentation task. Specifically, we randomly split original training set into 5 parts, where 4 parts are used for training while the rest one is used for validation. The best model on validation set is used for testing. Under this configuration, there may have overlapping samples (different scans from same patients) between training and validation sets. Strictly speaking, training and validation sets are supposed to be independent. Since the aim of cross-validation is to select the best-performed model on testing set, it is still reliable to select best-performed model on validation set with small number of random overlapping samples in training set.
\begin{table}
  \caption{Ablation experiments adapting OTU\_2d to OTU\_CEUS. ``Symmetric'' denotes the symmetric architecture with dual encoders and decoder-heads, i.e. DS$^2$Net removing $D_{E_{s}}$, $D_{E_{t}}$, DDSM and DUSM (Fig.\ref{Fig4}). ``+'' means appending.}
  \centering
  \resizebox{\linewidth}{!}{
  \begin{tabular}{c |c c c c|c c}
  \cmidrule(r){1-7}
   Methods &$\mathcal{L}_{seg}^{ss}$ &$\mathcal{L}_{seg}^{st}$ &$\mathcal{L}_{adv}^{E_{s}}$ &$\mathcal{L}_{adv}^{E_{t}}$ &$IoU (\%)$   &$mIoU (\%)$ \\
  \cmidrule(r){1-7}
  PSPNet (w/o DA)  &-  &-  &-  &-  & 50.22 & 67.60 \\
  \cmidrule(r){1-7}
  + Symmetric (w/o DA)   &\checkmark  &\checkmark &-  &- & 50.54 & 67.81 \\
  \cmidrule(r){1-7}
  + Feature alignment (FA)   &\checkmark  &\checkmark &\checkmark  &\checkmark &57.08 &72.55 \\
  \cmidrule(r){1-7}
  + DDSM   &\checkmark  &\checkmark &\checkmark  &\checkmark &59.28 &74.53 \\
  \cmidrule(r){1-7}
  + DUSM (DS$^2$Net\_P)  &\checkmark  &\checkmark &\checkmark  &\checkmark &61.85 &75.75 \\
  \cmidrule(r){1-7}
  SegFormer (w/o DA)  &-  &-  &-  &-  &61.11 &75.45 \\
  \cmidrule(r){1-7}
  + Symmetric (w/o DA)   &\checkmark  &\checkmark &-  &- &61.01 &75.33 \\
  \cmidrule(r){1-7}
  + Feature alignment (FA)   &\checkmark  &\checkmark &\checkmark  &\checkmark &64.74 &77.54 \\
  \cmidrule(r){1-7}
  + DDSM   &\checkmark  &\checkmark &\checkmark  &\checkmark &68.06 &79.78 \\
  \cmidrule(r){1-7}
  + DUSM (DS$^2$Net\_T)  &\checkmark  &\checkmark &\checkmark  &\checkmark &69.81 &80.86 \\
  \cmidrule(r){1-7}
  \end{tabular}}
 \label{Tab8}
\end{table}
\par Tab.\ref{Tab7} shows the comparison results. Generally, our proposed DS$^2$Net performs much stronger than previous methods. Specifically, From experimental results shown in Tab.\ref{Tab7}, we analyze in the following points. (1) In recent year, Transformer structure has been proved that it can outperform CNN structure in various vision tasks. However, it is more unstable in training phrase. The results in Tab.\ref{Tab7} show that SegFormer and DS$^2$Net\_T respectively has less bias but larger variance than PSPNet and DS$^2$Net\_P. (2) When DS$^2$Net is trained on OTU\_CEUS and evaluated on OTU\_2d, DS$^2$Net has larger variance compared to model trained on OTU\_2d and evaluated on OTU\_CEUS. The reason is that OTU\_CEUS has far smaller number of training samples than OTU\_2d, which leads to larger performance fluctuation. (3) DDFSeg and EDRL apply feature disentanglement mainly for better image translation. DCLA designs domain translation path, which also a typical image alignment methods. On MMOTU image dataset with small domain shift and small training set, three methods suffer from large performance fluctuation. DAFormer achieves competitive results, which benefits from Transformer encoder and self-training strategy. However, on binary segmentation task, it is easy to overfit on foreground category. (4) AdapSegNet, EGUDA, DDFSeg, DAFormer, EDRL and DCLA are all generalized and notable methods on UDA medical semantic segmentation task. Compared to them, DS$^2$Net is more suitable for specific data distribution and UDA task configuration of our proposed MMOTU image dataset, DS$^2$Net performs best among these four methods.
\subsubsection{Ablation Study} To separately prove the efficiency of each module of DS$^2$Net, we conduct experiments adapting OTU\_2d to OTU\_CEUS. From Tab.~\ref{Tab8}, we analyze in the following points. (1) Compared to PSPNet without domain adaptation, only applying symmetric structure can not bring improvement, because domain shift problem still exists. (2) When adding feature alignment mechanism, the performance significantly improve. This reflects that feature alignment makes effect on alleviating the domain shift problem and improves the general representation capability of $E_{s}$ and $E_{t}$. (3) When appending DDSM, the architecture further improves. The reason is that DDSM guides the optimization trend making the target images become easier to be represented into target-style. (4) When appending DUSM, the architecture still achieves improvement benefiting from universal feature extraction. (5) As shown in Tab.\ref{Tab8}, DDSM and DUSM can work well singly. When applying two modules together, the cross-domain semantic segmentation performance further improves. It means two modules are compatible and have complementary contributions on performance improvement. (6) In this experiment, DS$^2$Net is trained on training set and directly evaluated on testing set at each 4k iterations (80k iterations in total). We find a best performance model on testing set and provide the best results to reveal the effectiveness of each key module.
\subsubsection{Visualization and Analysis} 
\begin{figure*}
\begin{center}
   \includegraphics[width=0.95\linewidth]{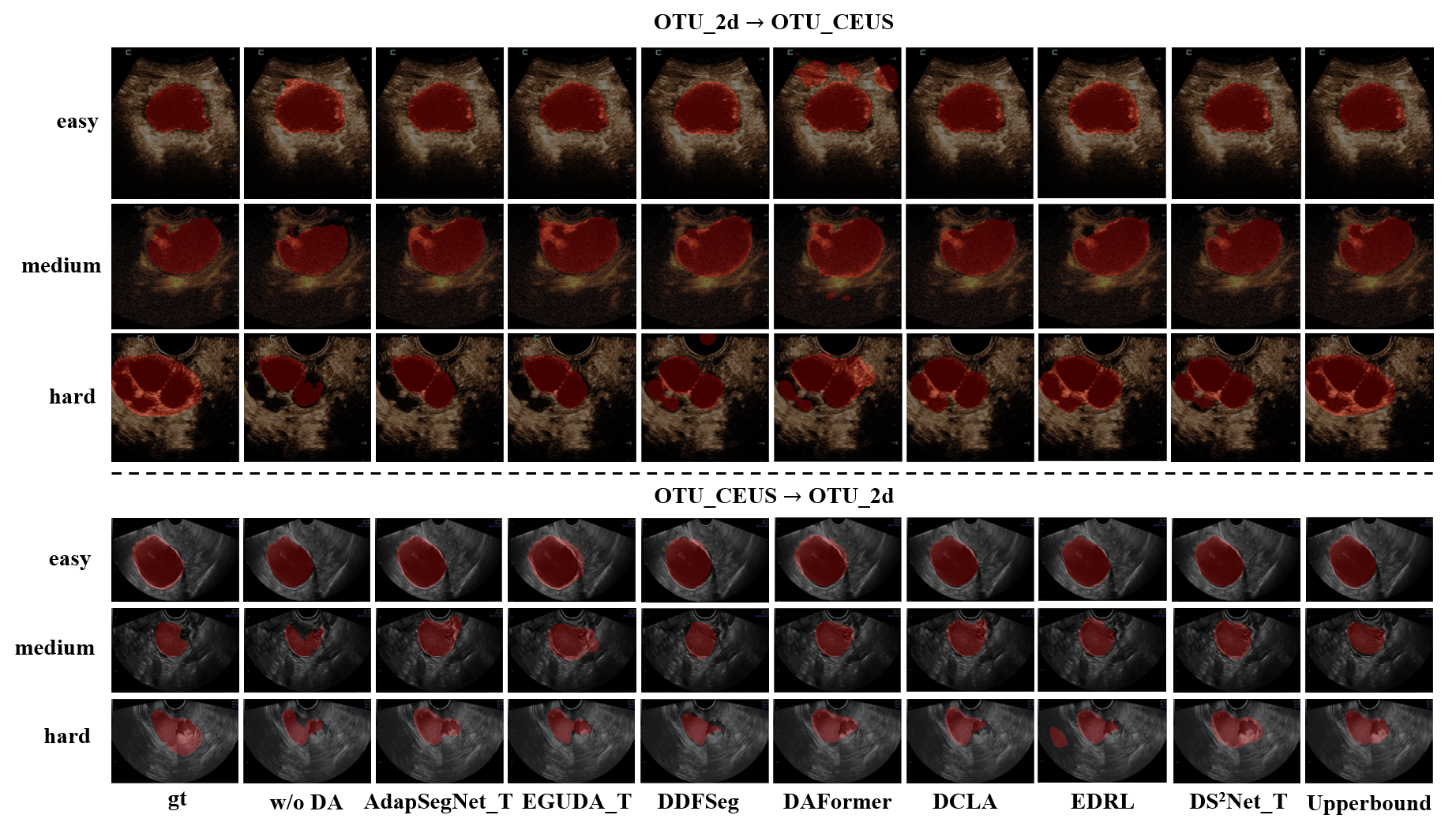}
\end{center}
   \caption{Visualization comparison between different methods. We select SegFormer-based models as example. ``Upperbound'' indicates the results from single-modality training model.}
\label{Fig7}
\end{figure*}

To intuitively show the performance of DS$^2$Net, we compare the segmentation performance of different methods. The qualitative results are shown in Fig.\ref{Fig7}. We can observe that training model under ``w/o DA'' setting can work well on some easy cases, but cannot handle harder cases. Benefiting from feature alignment mechanism, AdapSegNet and EGUDA can generate reasonable predictions under most cases, even on hard cases. DDFSeg and DAFormer can also perform well, in which DAFormer has more tendency on foreground prediction because of pseudo-foreground-label guidance (self-training mechanism). Obviously, DS$^2$Net shows the most powerful performance compared to previous methods. On medium and hard cases, our proposed DS$^2$Net can capture more semantic information and achieve closest results to ``Upperbound''. 
\begin{figure}[tb]
\begin{center}
   \includegraphics[width=1.0\linewidth]{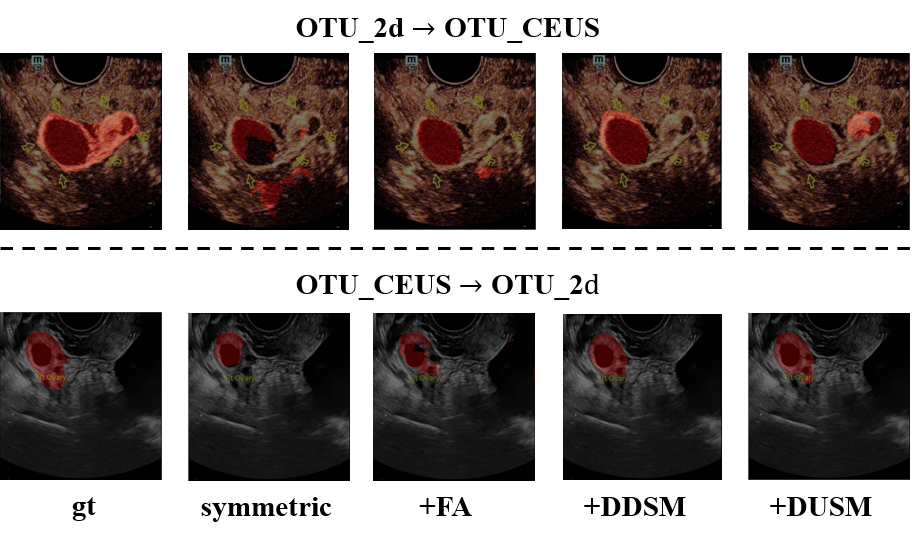}
\end{center}
   \caption{Visualization comparison to illustrate the effectiveness of each module. We select segformer-based models as example.}
\label{Fig8}
\end{figure}
\par To directly show the effectiveness of each module, we also conduct visualization analysis. As shown in Fig.\ref{Fig8}, symmetric architecture does not work well, because dual encoders and decoder-heads can not solve domain shift problem. When applying feature alignment mechanism, the model can generate reasonable predictions. When appending DDSM, the model shows stronger generalization power on target images. Intuitively, models with DDSM decrease obvious false positive predictive regions and avoid obvious missing detection. When adding DUSM, the final DS$^2$Net shows the strongest representation capability with better semantic understanding on target images, which can localize more ``hard-finding'' lesion regions. This visualization comparison coincides with the ablation experiments shown in Tab.\ref{Tab8}, which further proves the interpretability of DS$^2$Net in an intuitive manner.
\begin{table}
  \caption{Segmentation experiments on bidirectional CT-MR unsupervised domain adaptation. All methods are evaluated with Dice metric (\%). $^{\dag}$ indicates results reported from \cite{DDFSeg}.}
  \centering
  \resizebox{\linewidth}{!}{
  \begin{tabular}{c |c c c |c}
  \cmidrule(r){1-5}
  \multicolumn{5}{c}{CT $\rightarrow$ MR}
  \\
  \cmidrule(r){1-5}
   Methods & MYO & RV & LV & Mean 
   \\ \cmidrule(r){1-5}
   PSPNet (supervised) \cite{PSPNet} & 78.1$\pm$3.2 & 85.7$\pm$2.5 & 91.0$\pm$2.8 & 84.9$\pm$3.0
   \\ \cmidrule(r){1-5}
   SegFormer (supervised) \cite{SegFormer} & 84.3$\pm$4.1 & 90.7$\pm$3.2 & 92.6$\pm$3.8 & 89.2$\pm$4.1
   \\ \cmidrule(r){1-5}
   PSPNet (w/o DA) \cite{PSPNet} & 24.6$\pm$3.2 & 57.1$\pm$3.5 & 66.4$\pm$3.8 & 49.4$\pm$3.5
   \\ \cmidrule(r){1-5}
   SegFormer (w/o DA) \cite{SegFormer} & 40.8$\pm$4.6 & 68.1$\pm$4.0 & 73.9$\pm$4.8 & 60.9$\pm$4.9
   \\ \cmidrule(r){1-5}
   CycleSeg$^{\dag}$ \cite{cyclegan} & 53.2$\pm$17.1 & 79.2$\pm$13.1 & 81.3$\pm$11.8 & 71.2$\pm$19.1
   \\ \cmidrule(r){1-5}
   EGUDA\_T \cite{EGUDA} & 64.7$\pm$5.5 & 76.4$\pm$4.2 & 80.5$\pm$4.6 & 73.9$\pm$5.0
   \\ \cmidrule(r){1-5}
   DAFormer \cite{DAFormer} & 66.4$\pm$3.7 & 82.8$\pm$4.9 & 85.0$\pm$4.4 & 78.1$\pm$4.5
   \\ \cmidrule(r){1-5}
   SIFA$^{\dag}$ \cite{SIFA} & 67.3$\pm$11.4 & 84.2$\pm$11.5 & 87.6$\pm$8.9 & 79.6$\pm$13.9
   \\ \cmidrule(r){1-5}
   DDFSeg$^{\dag}$ \cite{DDFSeg} & 71.3$\pm$10.6 & 83.2$\pm$11.7 & 87.7$\pm$10.4 & 80.7$\pm$12.9
   \\ \cmidrule(r){1-5}
   EDRL \cite{EDRL} & \textbf{75.6$\pm$6.9} & \textbf{87.1$\pm$4.7} & \textbf{91.6$\pm$5.7} & \textbf{84.8$\pm$4.1}
   \\ \cmidrule(r){1-5}
   DS$^2$Net\_P & 63.5$\pm$4.0 & 74.3$\pm$3.7 & 80.4$\pm$4.4 & 72.7$\pm$4.3
   \\ \cmidrule(r){1-5}
   DS$^2$Net\_T & 70.2$\pm$4.3 & 82.5$\pm$5.0 & 87.0$\pm$4.1 & 79.9$\pm$5.1
   \\ \cmidrule(r){1-5}
   \multicolumn{5}{c}{MR $\rightarrow$ CT}
   \\ \cmidrule(r){1-5}
   PSPNet (supervised) \cite{PSPNet} & 80.9$\pm$2.8 & 86.1$\pm$3.0 & 89.6$\pm$2.5 & 85.5$\pm$3.0
   \\ \cmidrule(r){1-5}
   SegFormer (supervised) \cite{SegFormer} & 88.4$\pm$3.2 & 92.1$\pm$3.2 & 93.3$\pm$3.8 & 91.3$\pm$3.5
   \\ \cmidrule(r){1-5}
   PSPNet (w/o DA) \cite{PSPNet} & 27.8$\pm$2.2 & 48.8$\pm$2.9 & 58.3$\pm$2.4 & 45.0$\pm$2.8
   \\ \cmidrule(r){1-5}
   SegFormer (w/o DA) \cite{SegFormer} & 38.1$\pm$4.6 & 65.4$\pm$4.3 & 74.0$\pm$3.8 & 59.1$\pm$4.4
   \\ \cmidrule(r){1-5}
   EGUDA\_T \cite{EGUDA} & 52.2$\pm$6.3 & 77.4$\pm$4.3 & 81.9$\pm$5.2 & 70.5$\pm$5.7
   \\ \cmidrule(r){1-5}
    CycleSeg$^{\dag}$ \cite{cyclegan} & 51.3$\pm$15.4 & 83.3$\pm$7.7 & 79.3$\pm$15.3 & 71.3$\pm$19.5
   \\ \cmidrule(r){1-5}
   SIFA$^{\dag}$ \cite{SIFA} & 56.6$\pm$12.4 & 80.0$\pm$8.3 & 82.6$\pm$12.6 & 73.1$\pm$16.3
   \\ \cmidrule(r){1-5}
   DAFormer \cite{DAFormer} & 61.8$\pm$5.1 & 78.3$\pm$3.2 & 81.7$\pm$4.9 & 73.9$\pm$5.4
   \\ \cmidrule(r){1-5}
   DDFSeg$^{\dag}$ \cite{DDFSeg} & 66.9$\pm$11.0 & 79.1$\pm$6.7 & 83.5$\pm$16.0 & 76.5$\pm$13.8
   \\ \cmidrule(r){1-5}
   EDRL \cite{EDRL} & \textbf{77.1$\pm$11.1} & \textbf{87.0$\pm$9.1} & \textbf{87.4$\pm$16.6} & \textbf{84.0$\pm$10.3}
   \\ \cmidrule(r){1-5}
   DS$^2$Net\_P & 53.6$\pm$4.7 & 72.0$\pm$4.2 & 82.3$\pm$5.0 & 69.3$\pm$4.9
   \\ \cmidrule(r){1-5}
   DS$^2$Net\_T & 65.4$\pm$5.7 & 78.6$\pm$4.6 & 85.7$\pm$5.8 & 76.6$\pm$6.0
   \\ \cmidrule(r){1-5}
  \end{tabular}}
 \label{Tab_CTMR}
\end{table}
\begin{figure}[tb]
\begin{center}
   \includegraphics[width=1.0\linewidth]{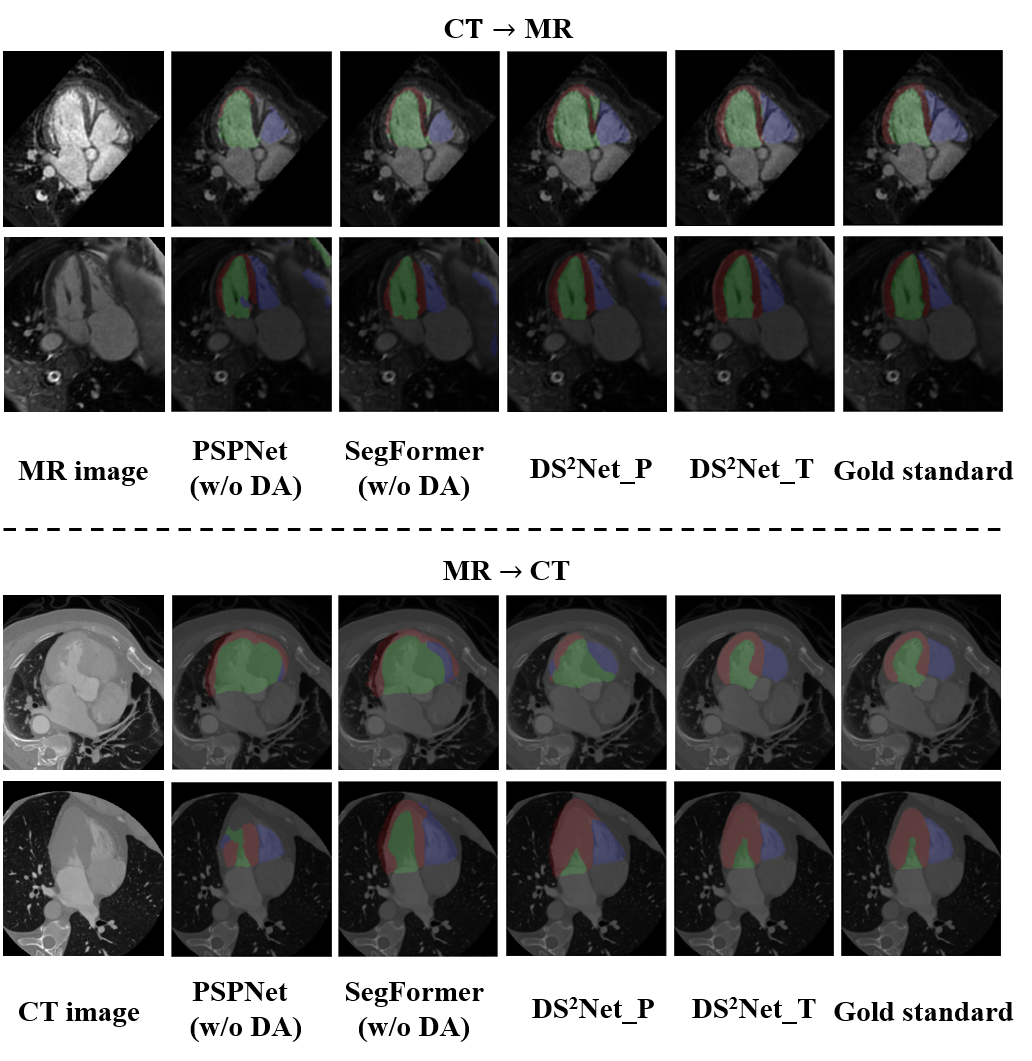}
\end{center}
   \caption{Visualization of segmentation results on bidirectional CT-MR unsupervised domain adaptation. The region of MYO, RV and LV are respectively indicated by red, blue and green. Here, electronic version are recommended for readers.}
\label{Fig11}
\end{figure}
\subsubsection{Generalization Analysis} To demonstrate the generalization of our proposed DS$^2$Net, we further conduct experiments on MM-WHS (Multi-Modality Whole Heart Segmentation)~\cite{MM-WHS1, MM-WHS2}, which is a notable cross-modality medical image dataset including CT and MR images\footnote{https://github.com/FupingWu90/CT\_MR\_2D\_Dataset\_DA}. This dataset consists of 52 CT images and 46 MR images, where 20 CT and 20 MR are labeled with gold standard segmentation mask of LV, MYO and RV~\cite{MM-WHS1, MM-WHS2}. For experiments, we use the 16 slices from each 3D image. These 2D slices were extracted from the long-axis view around the center of left ventricular cavity~\cite{MM-WHS2}. Specifically, 320 labeled and 512 unlabeled slices are from CT images, while 320 labeled and 416 unlabeled slices are from MR images.
\par To evaluate the generalization of DS$^2$Net, we follow~\cite{DDFSeg} and conduct cross-domain experiments (5-Fold cross-validation) including adaptation from CT images to MR image and adaptation from MR images to CT image. The segmentation results (Dice metric) and visualization results are respectively shown in Tab.\ref{Tab_CTMR} and Fig.\ref{Fig11}. Here, we'd like to analyze in the following points. (1) Compared to DDFSeg, DS$^2$Net\_T achieves comparable results. On ``CT $\rightarrow$ MR'' UDA segmentation task, DS$^2$Net\_T has little inferiority than DDFSeg (0.8\% less in Dice metric). On ``MR $\rightarrow$ CT'' UDA segmentation task, DS$^2$Net\_T outperforms DDFSeg by 0.1\% in Dice metric. (2) Obviously, EDRL is the currecent SOTA methods. Even though DS$^2$Net is inferior to EDRL, the second-best results can still prove its generalization capability. (3) Obviously, DS$^2$Net has less standard deviation than previous methods. The main reason is that CycleSeg, SIFA, DDFSeg and EDRL all adopt pixel-to-pixel image translation for alignment. Compared to image alignment mechanism, feature alignment mechanism is more stable during optimization. Our proposed DS$^2$Net is a typical feature alignment based method, which leads to low-fluctuation among different runs. (4) Compared to OTU\_2d/OTU\_CEUS, CT/MR images obviously have larger domain shift. In this case, image alignment based methods have more advantages. (5) In Fig.\ref{Fig11}, we provide 4 cases to visualize the segmentation prediction on bidirectional CT-MR unsupervised domain adaptation. This visualization coincides with results shown in Tab.\ref{Tab_CTMR}. As a whole, experimental results on CT/MR (MM-WHS) UDA semantic segmentation task prove the generalization of DS$^2$Net.
\begin{table}
  \caption{Single-modality recognition results on MMOTU image dataset. We adopt top1 and top2 accuracy as our metric.}
  \centering
  \scalebox{0.85}{
  \begin{tabular}{c c c c c}
  \cmidrule(r){1-5}
  \multirow{2}{*}{Methods}    &\multicolumn{2}{c}{OTU\_2d}        &\multicolumn{2}{c}{OTU\_CEUS}   \\
  \cmidrule(r){2-5}
                         & top1 (\%)   & top2 (\%)
                         & top1 (\%)   & top2 (\%)      \\
  \cmidrule(r){1-5}
  VGG-16  & 67.59  & 79.96  & 61.0  & 74.0 \\
  \cmidrule(r){1-5}
  ResNet-34  &77.61  &88.70  &69.0  &79.0 \\
  \cmidrule(r){1-5}
  ResNet-50  &80.17  &90.19  &72.0  &81.0 \\
  \cmidrule(r){1-5}
  DenseNet-121    &78.89  &89.34  &69.0  &77.0 \\
  \cmidrule(r){1-5}
  MobileNetV2  & 76.97  & 86.78 & 68.0  & 78.0 \\
  \cmidrule(r){1-5}
  EfficientNet-b0 &76.12  &85.93 &64.0  &79.0  \\
  \cmidrule(r){1-5}
  EfficientNet-b1 &77.61 &86.99 &71.0 &79.0 \\
  \cmidrule(r){1-5}
  EfficientNetV2-S  & 79.74  & 89.55 & 70.0  & 80.0 \\
  \cmidrule(r){1-5}
  EfficientNetV2-M  & 80.60  & 91.04 & 71.0  & 81.0 \\
  \cmidrule(r){1-5}
  \end{tabular}}
 \label{Tab9}
\end{table}
\subsection{Single-Modality Recognition}
\par On Task1 and Task2, we apply a ``foreground-background'' lesion area segmentation without directly applying seven/eight-category segmentation. Therefore, we tackle single-modality image recognition in Task3 (Fig.\ref{Fig1}) to classify ovarian tumors. In this paper, we provide VGG~\cite{VGG}, ResNet~\cite{ResNet}, DenseNet~\cite{densenet}, EfficientNet~\cite{EfficientNet}, MobileNetV2~\cite{MV2} and EfficientNetV2~\cite{EfficientNetV2} as baseline classification models. For all models, we select SGD as optimizer. The momentum and weight decay value are set as 0.9 and 0.0005. All model are loaded ImageNet-pretrained weights before training. As shown in Tab.\ref{Tab9}, EfficientNetV2-M outperforms other methods on OTU\_2d while ResNet-50 achieves close results. On OTU\_CEUS, ResNet-50 outperforms other methods while EfficientNet-b1 and EfficientNetV2-M achieve close results. Generally, ResNet-50 and EfficientNetV2 show strong generalization capability on single-modality recognition task. Moreover, besides top1 accuracy, we also use top2 accuracy as metric because providing two category candidates is also valuable for diagnosis in practical medical treatment.
\begin{figure}[tb]
\begin{center}
   \includegraphics[width=0.95\linewidth]{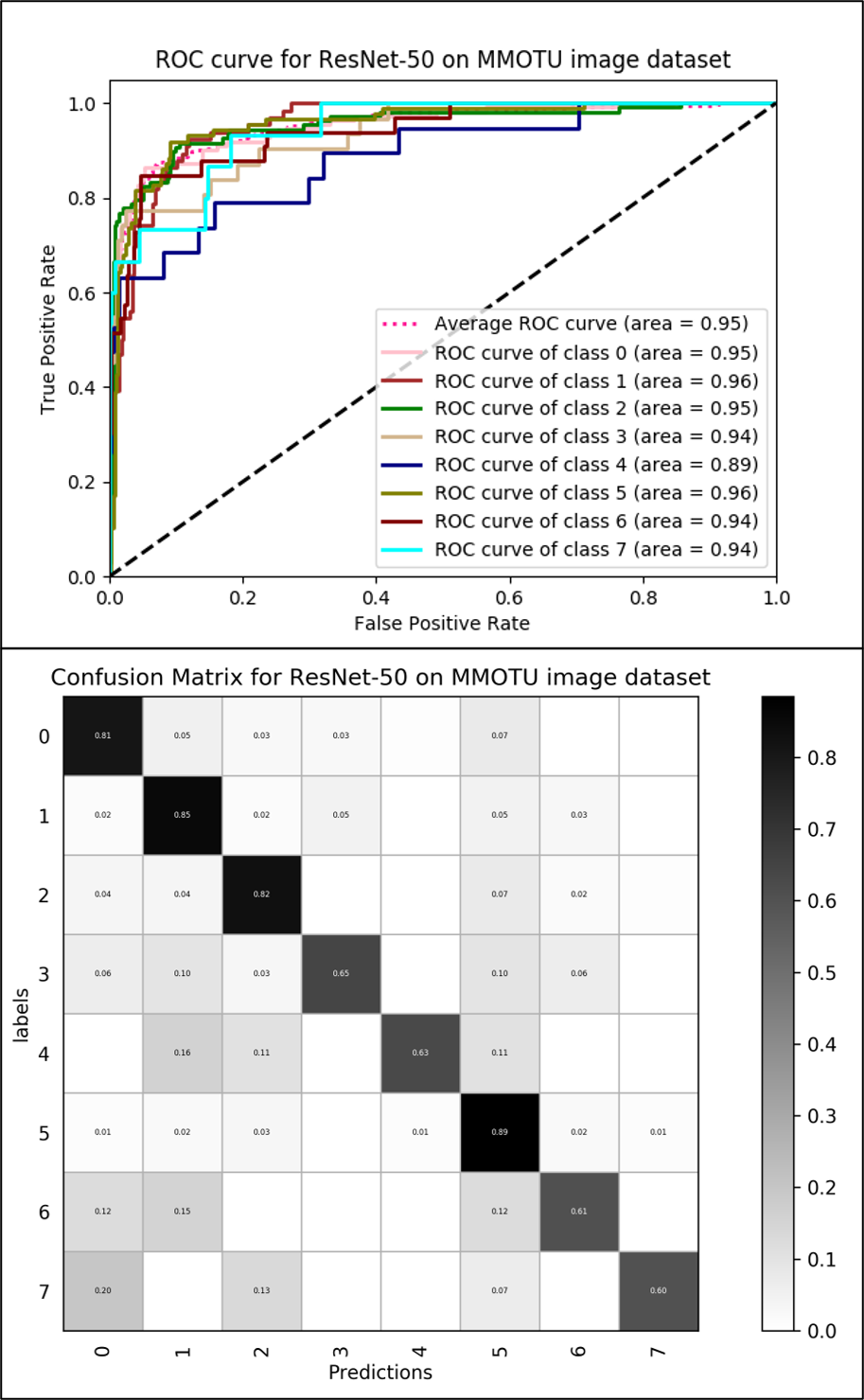}
\end{center}
   \caption{ROC curve and confusion matrix on 8 ovarian tumor categories. Here, we conduct experiments on OTU\_2d and select ResNet-50 as classification network.}
\label{Fig9}
\end{figure}
\par To better present the single-modality recognition performance, we select ResNet-50 as classification network and further evaluate it with ROC curve and confusion matrix on OTU\_2d. Obviously, experiments results from classification accuracy (Tab.\ref{Tab9}), AUC (area under curve) and confusion matrix are mutually corroborated. We also find that the Chocolate cyst, Serous cystadenoma, Teratoma and Normal ovary (category 0, 1, 2, 5) obviously have better classification accuracy than Theca cell tumor, Simple cyst, Mucinous cystadenoma and High grade serous (category 3, 4, 6, 7). The main reason is that ``category 0, 1, 2, 5'' have more training samples the ``category 3, 4, 6, 7'' (Fig.\ref{Fig_2}).
\subsection{Discussion}
\par In this paper, we propose MMOTU image dataset containing 2d ultrasound and CEUS images. Based on our proposed dataset, we tackle three tasks and mainly focus on bidirectional UDA segmentation between 2d ultrasound and CEUS images. To boost the cross-domain segmentation performance, we propose DS$^2$Net, which works well on closing the representation gap between 2d ultrasound and CEUS images. Our research provides an insight on detecting ovarian tumors on multi-modality ultrasound images, especially on CEUS images. Our research can also be regarded as an AI-aided technique for clinical diagnosis and treatment.
\subsubsection{Discussion on the Risk of Model Biasing by Symbols on Images}
\par As mentioned in Fig.\ref{Fig2}, there are some symbols on images. Due to historical and practical reasons, we can not collect pure raw data. Naturally, there may raise a question, is there a risk of biasing the model by symbols during optimization? In this section, we conduct experiments on single-modality 2d ultrasound image segmentation and cross-domain segmentation tasks to analyze the influence of symbols.
\par To remove the symbols from images, we apply the recent notable image inpainting method~\cite{deepfillv2} on all samples in MMOTU image dataset. Then we use the original and ``symbol-free'' MMOTU image dataset to respectively train models. We will discuss the results shown in Tab.\ref{Tab11} as follows. (1) Generally, models trained with ``symbol-free'' samples show performance decline. (2) This comparison is a little bit unreasonable because the training and testing set both changes after removing the symbols. However, we can still roughly draw a conclusion that applying image inpainting technique is not necessary. (3) Intuitively, images after removing symbols are not becoming distorted or destroyed. However, it also change the data in an implicit manner. It means that the negative influence from data distribution change outweighs the positive influence from symbol interference. (4) From another perspective, experts are not confused by those symbols during annotating the images, which means the symbols have little impact on visual features of images. Therefore, we believe that our models can also overcome the interference of symbols and learn those important and useful features. 
\begin{table}
  \caption{Semantic segmentation results comparison between models trained by images with and w/o symbols. On single-modality segmentation, we select PSPNet and Segformer as example.}
  \centering
  \resizebox{\linewidth}{!}{
  \begin{tabular}{c c c c c}
  \cmidrule(r){1-5}
  \multicolumn{5}{c}{\textbf{Single-modality segmentation results on MMOTU image dataset}} \\
  \cmidrule(r){1-5}
  \multirow{2}{*}{Methods}    &\multicolumn{2}{c}{OTU\_2d}        &\multicolumn{2}{c}{OTU\_CEUS}   \\
  \cmidrule(r){2-5}
                         &$IoU (\%)$   &$mIoU (\%)$
                         &$IoU (\%)$   &$mIoU (\%)$      \\
  \cmidrule(r){1-5}
  PSPNet &82.01  &89.41  &71.01  &81.6 \\
  \cmidrule(r){1-5}
  PSPNet (w/o symbols) &81.13  &88.63  &70.78  &81.09 \\
  \cmidrule(r){1-5}
  SegFormer &82.46   &89.88  &73.03  &83.0  \\
  \cmidrule(r){1-5}
  SegFormer (w/o symbols) &81.52   &89.17  &72.62  &82.55  \\
  \cmidrule(r){1-5}
  \multicolumn{5}{c}{\textbf{UDA segmentation results on MMOTU image dataset}} \\
  \cmidrule(r){1-5}
  \multirow{2}{*}{Methods}    &\multicolumn{2}{c}{OTU\_CEUS $\rightarrow$ OTU\_2d}        &\multicolumn{2}{c}{OTU\_2d $\rightarrow$ OTU\_CEUS}   \\
  \cmidrule(r){2-5}
                         &$IoU (\%)$   &$mIoU (\%)$
                         &$IoU (\%)$   &$mIoU (\%)$      \\
  \cmidrule(r){1-5}
  DS$^2$Net\_P &54.06 &71.87 &61.85 &75.75 \\
  \cmidrule(r){1-5}
  DS$^2$Net\_P (w/o symbols) &52.93 &70.46 &60.49 &74.62 \\
  \cmidrule(r){1-5}
  DS$^2$Net\_T &65.42 &79.20 &69.81 &80.86 \\
  \cmidrule(r){1-5}
  DS$^2$Net\_T (w/o symbols) &64.03 &78.02 &68.58 &79.43 \\
  \cmidrule(r){1-5}
  \end{tabular}}
 \label{Tab11}
\end{table}
\par As a whole, symbols on images may harm segmentation performance, but we find the harm is limited from the experiments. Moreover, inpainting on MMOTU image dataset is not recommended. It'd be better to preserve raw data.
\subsubsection{Limitations and Future Works} However, our research still has some limitations to be improved. Here, we summarize four main limitations as follows. (1) As shown in Fig.\ref{Fig2}, some symbols are left on our collected images. Due to some historical and practical reasons, we can not collect pure raw data without symbols, which may cause model biasing by symbols during optimization. (2) MMOTU image dataset has only 170 CEUS images. Lacking both training and testing images will harm the credibility of experiments. (3) DS$^2$Net shows strong performance on CEUS images (single-frame) trained with 2d ultrasound images. However, DS$^2$Net can not directly apply on CEUS sequences, which means there still exists challenges to spread our proposed AI-aided technique on clinical CEUS examination. (4) Our paper lacks of detailed analysis on recognition task (Task3, Fig.\ref{Fig1}). Although we mainly focus on lesion segmentation, recognition of tumor type is also a meaningful task in practical clinical diagnosis.
\par In the future, we will further extend our work and focus in the following aspects. (1) We will continually extend our proposed MMOTU image dataset, especially OTU\_CEUS. We will also pay attention to preserving pure data without symbols. (2) Actually, the total tumor categories of OTU\_2d are 32. Since some of tumor types (e.g. fibrosarcoma) have only one or two samples, we exclude them in our dataset. In the future, we will extend the categories together with collecting more data. (3) In MMOTU image dataset, samples of each category are unbalanced. We will further explore its influence. (4) In this work, we explore the AI-aided techniques on CEUS images. In the future, we will further explore the AI-aided methods on CEUS sequences, which is more valuable and meaningful. (5) We will try to migrate our method on other cross-modality data (e.g., CT/MRI) for cross-domain adaptation segmentation task to prove the robustness of our method.
\section{Conclusion}
\label{sec:conclusion}
In this paper, we propose a multi-modality ovarian tumor ultrasound (MMOTU) image dataset to explore the cross-domain representation potential. MMOTU image dataset contains 1469 2d ultrasound images and 170 CEUS images with pixel-wise and global-wise annotations. Based on MMOTU image dataset, we mainly focus on alleviating the domain shift problem on lesion area segmentation for bidirectional unsupervised domain adaptation between 2d ultrasound and CEUS images. In this paper, we propose DS$^2$Net, which is a feature alignment based architecture. Specifically, we use adversarial learning to first ease the domain shift of encoders for better representation on both source and target images. Then, we design DDSM and DUSM to extract the domain-distinct and domain universal features. With these two modules, each encoder and decoder-head will further have specific style feature representation capability towards both source and target images. Extensive experiments show that the DS$^2$Net outperforms the previous notable feature alignment based methods on MMOTU image dataset. Visualization and analysis also prove that DS$^2$Net is convincing and interpretable. In this paper, we also provide series of baseline models on single-modality semantic segmentation and recognition tasks. Compared to previous methods, we first propose method to tackle the cross-domain ovarian tumor segmentation on our proposed MMOTU image dataset, which provides a new insight on detecting ovarian tumors. 

\section*{CRediT Author Statement}
\textbf{Qi Zhao}: Conceptualization, Supervision, Project administration. \textbf{Shuchang Lyu}: Conceptualization, Methodology, Software, Writing - Original Draft. \textbf{Wenpei Bai}: Validation, Writing - Review \& Editing, Funding acquisition. \textbf{Linghan Cai}: Data Curation, Formal analysis. \textbf{Binghao Liu}: Validation, Visualization. \textbf{Guangliang Cheng}: Supervision, Writing - Review \& Editing. \textbf{Meijing Wu}: Data Curation. \textbf{Xiubo Sang}: Data Curation. \textbf{Min Yang}: Resources. \textbf{Lijiang Chen}: Investigation, Writing - Review \& Editing.

\section*{Acknowledgment}
\label{sec:acknowledgment}
All the data collected in our proposed dataset have appropriate approvals from the ethics committees of Beijing Shijitan Hospital, Capital Medical University. This work was supported by the National Natural Science Foundation of China (grant numbers 62072021).

\bibliographystyle{cas-model2-names}

\bibliography{cas-refs}

\end{document}